\documentclass[article,12pt,authoryear,preprint]{elsarticle}

\usepackage{amssymb}
\usepackage{multirow}
\usepackage{amsmath}
\usepackage{amsfonts}
\usepackage{booktabs}
\usepackage{tabularx}
\usepackage{array}
\usepackage{float}
\newcommand{\E}{\mathbb{E}}
\usepackage{bbm}
\usepackage{tikz}
\usetikzlibrary{arrows.meta, positioning, calc, shapes.geometric}
\usepackage{hyperref}
\usepackage{graphicx}
\usepackage{xcolor}
\newcommand{\N}{\mathbb{N}}
\usepackage{bm}
\usepackage{algorithm}
\usepackage{algpseudocode}
\definecolor{deficitc}{RGB}{200, 80, 80}
\definecolor{deficitlc}{RGB}{240, 200, 200}
\definecolor{skyc}{RGB}{80, 150, 200}
\definecolor{skylc}{RGB}{200, 225, 240}
\definecolor{skymc}{RGB}{150, 200, 230}
\definecolor{sagec}{RGB}{100, 160, 130}
\definecolor{sagelc}{RGB}{200, 230, 210}
\journal{Computers \& Industrial Engineering}

\begin{document}
\onecolumn
\begin{frontmatter}

\title{CoRe-MARL: \textbf{Co}operative \textbf{Re}distribution Under Unknown Dynamics Using Recurrent Multi-Agent Reinforcement Learning}

\author[1]{Naimur Rahman Chowdhury\corref{cor1}}
\ead{nchowdh2@ncsu.edu}

\author[2]{Shatabdi Sen Prapti}
\ead{shatabdisprapti@gmail.com}

\author[2]{Md. Salehin Seyam}
\ead{2208018@ipe.buet.ac.bd}

\author[2]{Limon Bin Hossain}
\ead{mdlimonbinhossain@gmail.com}

\affiliation[1]{organization={This work was done prior to joining Amazon. Industrial and Systems Engineering, North Carolina State University},
                city={Raleigh},
                state={NC},
                country={United States}}

\affiliation[2]{organization={Department of Industrial and Production Engineering,
                              Bangladesh University of Engineering and Technology},
                city={Dhaka},
                postcode={1000},
                country={Bangladesh}}
\cortext[cor1]{Corresponding author}
\begin{abstract}
\noindent
Relief distribution programs operate through networks of facilities that face uncertain local demand and supply, so the availability of relief differs across the communities they serve. Redistributing supplies among centers can reduce these imbalances, but the centers often make this decision with limited information. This study develops CoRe-MARL, a cooperative multi-agent reinforcement learning (MARL) framework that formulates redistribution as a decentralized partially observable Markov decision process (Dec-POMDP). Each center is an agent that allocates its available perishable emergency supplies among local service, reserve, and redistribution to the other centers. We formulate a shared reward that favors the worst-served center and penalizes service gaps across centers, with an additional penalty when network-wide service falls below a threshold. To understand dynamics, we formulate a recurrent network that summarizes the evolving disruption, which the agents never observe directly, and multi-agent proximal policy optimization (MAPPO) provides centralized training and decentralized execution (CTDE). We evaluate recurrent MAPPO against recurrent independent PPO (IPPO) and rule-based heuristics on synthetic disruptions and a data-informed Hurricane Ian study across six southwest Florida counties. Our results show that, on average, MAPPO attains the smallest service gap and the highest worst-center service, and it outperforms IPPO in both network-wide and worst-center service. Learned cooperation therefore adds value over simple rules when protecting the worst-served communities is the priority, particularly when disruptions escalate.
\end{abstract}


\end{frontmatter}
\onecolumn

\section{Introduction}
\label{sec:intro}

Emergency assistance organizations, such as the Federal Emergency Management Agency (FEMA) and the American Red Cross, receive relief supplies and aim to distribute them to communities in need \citep{egan2010national}. The timely distribution of emergency supplies, such as food, water, and medicine, is critical to meeting the immediate needs of these vulnerable communities. However, these assistance programs usually operate in a network of distribution centers to get critical supplies closer to communities in times of need \citep{OzdamarErtem2015}. The supplies arrive at distribution centers on an ad hoc basis, driven by the timing and origin rather than the spatial distribution of need. Moreover, road damage, traffic disruptions, and communication failures further degrade the predictability of inbound supplies. Hence, a center's receipt of supply in any given period is only weakly correlated with the need in the surrounding affected regions \citep{Barbarosogluetal2002}. Due to the ad hoc nature of incoming supplies, some relief centers receive surpluses while others face acute shortages, even when the network as a whole might hold sufficient relief inventory. This imbalance is consequential, especially for perishable items. For perishable items, surplus stock that cannot be used locally within its shelf life may go to waste, which could otherwise be used to serve a center with insufficient supplies. Redistribution, the transfer of resources between distribution centers after their initial allocation, is essential in this regard to serve areas that need immediate support from centers with surplus supplies.

The existing literature on redistribution traditionally treats the problem as centralized planning. \citet{Rottkemperetal2011,Rottkemperetal2012} develop models for optimal inventory redistribution that address temporal changes in demand, and assume a single decision-maker with complete knowledge of the system. \citet{PachecoBatta2016} also model the prepositioning of inventories based on hurricane forecasts from a central perspective. These approaches demonstrate that redistribution can improve outcomes when a planner has full visibility. However, during emergencies, distribution centers operate with limited information about the rest of the network, relying on delayed, aggregated reports of other sites' operations \citep{ye2020,balcik2010}. Centralized optimization, therefore, struggles to capture the day-to-day decisions of the relief centers.

On the other hand, cooperation among centers is necessary to mitigate the imbalance in the network. For instance, a center with a surplus is unaware of which partner is most in need, and a center facing a shortage would not know which partner has stock to redistribute unless the centers communicate and coordinate their actions. In addition, the emergency condition, such as a natural disaster, itself evolves through phases that differ across locations and over time. Some communities experience a sudden peak, others face a prolonged disruption. These temporal trajectories are not known to decision-makers at centers in advance, and they must make decisions based on the limited observations they receive. A redistribution policy should therefore incorporate cooperation to correct imbalances, the partial observability of the network state, and the shifting dynamics that change a region's states.

In this study, we address the aforementioned requirements by formulating the relief redistribution problem as a cooperative multi-agent task. The centers are modeled as independent agents in a fully cooperative, partially observable, decentralized Markov decision process \citep{bernstein2002,oliehoek2016}. Each agent decides how to allocate its available supply among local service, holding for reserve, and transferring to the other centers. To handle the temporal structure of different dynamics, each agent also maintains a recurrent belief state that summarizes its history of observations, allowing it to infer the current phase of the emergency event and act accordingly. 

We formulate the following research questions.

\begin{itemize}
\item[\textbf{RQ1.}] \textit{How should a relief distribution center allocate its available supply between local service, holding for reserve, and redistribution to other centers under limited information about other centers and unknown dynamics?} 

To address this, we formulate the redistribution problem as a cooperative partially observable Markov game and solve it using recurrent multi-agent proximal policy optimization (MAPPO). Each center uses its local observation history and limited network information to dynamically allocate available inventory among local service, reserve, and redistribution. The recurrent policy enables agents (centers) to adapt their actions to evolving conditions without explicit knowledge of the trajectory.

\item[\textbf{RQ2.}] \textit{Can distribution centers coordinate redistribution decisions to improve service for the worst-served communities, and how does this equity gain trade off against overall network service?}

To answer this, we define a shared reward that penalizes dispersion of service across centers while rewarding the minimum service level, and evaluate the resulting trade-off across different supply and demand realizations. We compare the learned policies with local-only operation and with two decentralized surplus-sharing rules that require no training, so that the value of learning is measured against simple rules that a relief network could adopt immediately.

\item[\textbf{RQ3.}] \textit{Does a centralized cooperative learning help the network make better decisions compared with independent learning for relief redistribution?}

To answer this, we compare recurrent MAPPO with recurrent independent PPO (IPPO) on the same trajectories to assess how centralized training affects the action performance.
\end{itemize}

The remainder of the paper is organized as follows. Section \ref{sec:literature} reviews the related literature on humanitarian logistics, reinforcement learning in emergency response, and multi-agent coordination. Section \ref{sec:method} presents the mathematical formulation of the redistribution problem. Section \ref{sec:exp} describes the computational study: the relief network, the synthetic disruptions, the Hurricane Ian case, the baselines, and the evaluation protocol. Section \ref{sec:results} reports the results on held-out simulated disruptions and on a case study of Hurricane Ian in Southwest Florida (2022), built from the points of distribution operated in six counties, observed power outages and FEMA assistance registrations, and Section \ref{sec:managerial} derives managerial insights. Section \ref{sec:conclusion} concludes with a discussion of limitations and directions for future work.

\section{Literature Review}
\label{sec:literature}
Studies addressing resource allocation in emergency events focus on different aspects of logistics management, including efficiency of the balanced allocation and costs \citep{Sakianietal2020}, adequacy of the supply to demand realization under uncertainty \citep{Rottkemperetal2011,PachecoBatta2016}, and, most recently, equity of the outcome \citep{YuZhangetal2021}. In contrast to the majority of the literature, \citet{GutjahrFischer2018} consider equity explicitly as the objective of resource allocation decisions. The authors' contribution is to show that the minimal total deprivation cost function alone cannot adequately account for the inequity in deprivation costs and to propose adjusting it by explicitly incorporating the Gini coefficient of deprivation costs as a corrective measure.

Specifically, regarding resource redistribution, \citet{Rottkemperetal2012} develop a model for optimal inventory redistribution between exchange centers that accounts for changes in supply and demand following a disaster. \citet{PachecoBatta2016} propose a forecast-based approach that employs pre-positioned supplies, which are updated as the hurricane track becomes clearer. \citet{Sakianietal2020} propose a combined vehicle routing and network flow model, which they formulate as a rolling-horizon inventory routing problem with an objective function based on deprivation costs that embodies equity considerations. Our paper is related to the aforementioned articles in that it also concerns the redistribution during an emergency event with unknown dynamics. However, we study a decentralized setting in which each center makes distribution decisions on its own, based solely on local observations, with limited knowledge of the other centers’ stocks and demands.

Several works explore the application of RL to humanitarian logistics, primarily in single-agent settings. \citet{YuZhangetal2021} apply Q-learning to the allocation of relief resources with costs in efficiency, effectiveness, and equity estimated separately. \citet{LeeLee2021} frame the disaster response as a partially observable multi-agent task, but do not account for inventory perishability and multi-day delivery uncertainty. \citet{WuTai2024} combine convolutional networks with RL to optimize inbound logistics of food banks. Furthermore, \citet{VanSteenbergenEtAl2023} apply RL to humanitarian relief distribution using trucks and UAVs under travel-time uncertainty, and \citet{AhmadEtAl2025} propose a deep RL approach that jointly targets efficiency, effectiveness, and equity in disaster relief distribution, though both retain a single centralized decision-maker. Similarly, in multi-agent settings, \citet{YangEtAl2024} apply multi-agent deep RL to post-hazard community recovery planning, although their setting focuses on restoration scheduling rather than physical resource redistribution. In contrast to these studies, we incorporate equity into the reward function as a penalty for service gaps across agents and explicitly model the worst-case service across sites, thereby driving cooperative redistribution decisions. Second, we consider decision-making autonomy distributed among the centers with centralized training and decentralized execution (CTDE). 

Several works focus on the coordination mechanism. Prior MARL studies address decentralized inventory management, evaluate several algorithms for decentralized inventory control, and find that MAPPO has an edge over independent learning variants. \citet{Liuetal2025} apply a heterogeneous agent version of PPO to multi-echelon inventory management and find that it reduces costs and the variance of order quantities (the bullwhip effect) compared to single-agent RL. We differ from the approaches mentioned by focusing not only on the multi-agent setting but also by incorporating a recurrent network into the learning model to capture unknown dynamics. We address transport uncertainty, disruption-dependent shipment losses, and inventory perishability factors.

In sum, the body of literature covers centralized optimization to decentralized reinforcement learning. However, the question of fully decentralized coordination for redistribution under unknown dynamics has received insufficient attention. Most works focus on either centrally coordinated dispatching or inventory management in decentralized supply chains. This is an important gap, as many real-world relief networks are decentralized, with individual centers making day-to-day distribution decisions with only local information. In Table \ref{tab:literature_comparison} we present the literature gap and the contribution of this study. 

\begin{table}[H]
\centering
\caption{Comparison of the current paper with related studies on 
         resource allocation and redistribution.}
\label{tab:literature_comparison}
\footnotesize
\setlength{\tabcolsep}{4pt}
\renewcommand{\arraystretch}{1.25}
\resizebox{\textwidth}{!}{%
\begin{tabular}{l c c c c c l l}
\toprule
\multirow{2}{*}{\textbf{Reference}}
  & \textbf{Re-}
  & \textbf{Perish-}
  & \multirow{2}{*}{\textbf{RL}}
  & \textbf{Multi-agent}
  & \textbf{Decentralized}
  & \multirow{2}{*}{\textbf{Objective function}}
  & \multirow{2}{*}{\textbf{Solution approach}} \\
  & \textbf{distribution}
  & \textbf{ability}
  &
  & \textbf{Cooperation}
  & \textbf{Execution}
  & & \\
\midrule

\citet{Rottkemperetal2011}
  & \checkmark & & & & &
  Relocation cost/coverage &
  Exact MILP \\

\citet{Rottkemperetal2012}
  & \checkmark & & & & &
  Transshipment cost &
  Exact MILP \\

\citet{PachecoBatta2016}
  & \checkmark & & & & &
  Prepositioning cost &
  Forecast-driven heuristic \\

\citet{Sakianietal2020}
  & \checkmark & & & & &
  Deprivation \& operating cost &
  Specialized SA \\

\citet{GutjahrFischer2018}
  & & & & & &
  Deprivation-cost equity &
  Exact and metaheuristic \\

\citet{YuZhangetal2021}
  & & & \checkmark & & &
  Efficiency, effectiveness, equity &
  RL \\

\citet{WuTai2024}
  & & \checkmark & \checkmark & & &
  Quality and storage management &
  CNN + RL \\

\citet{VanSteenbergenEtAl2023}
  & & & \checkmark & & &
  Delivery under travel-time uncertainty &
  RL \\

\citet{AhmadEtAl2025}
  & & & \checkmark & & &
  Efficiency, effectiveness, equity &
  RL \\

\citet{LeeLee2021}
  & & & \checkmark & \checkmark & \checkmark &
  Admission/diversion decisions &
  Multi-agent RL \\

\citet{YangEtAl2024}
  & & & \checkmark & \checkmark & \checkmark &
  Post-hazard recovery scheduling &
  Multi-agent RL \\

\citet{Mousaetal2024}
  & & & \checkmark & \checkmark & \checkmark &
  Inventory cost &
  Multi-agent RL  \\

\citet{Liuetal2025}
  & & & \checkmark & \checkmark & \checkmark &
  Cost and order variance &
  Heterogeneous-agent PPO \\

\midrule
\textbf{This study}
  & \checkmark & \checkmark & \checkmark & \checkmark & \checkmark &
  Network service, Service gap, Worst-case Service &
  Recurrent MAPPO (CTDE) \\

\bottomrule
\end{tabular}%
}
\end{table}
\section{Methodology}
\label{sec:method}
In this section, we formulate the relief problem and develop the decentralized decision model. 
\subsection{Problem Formulation}
\label{sec:method-network}
\begin{figure}
    \centering
    \includegraphics[width=1\linewidth]{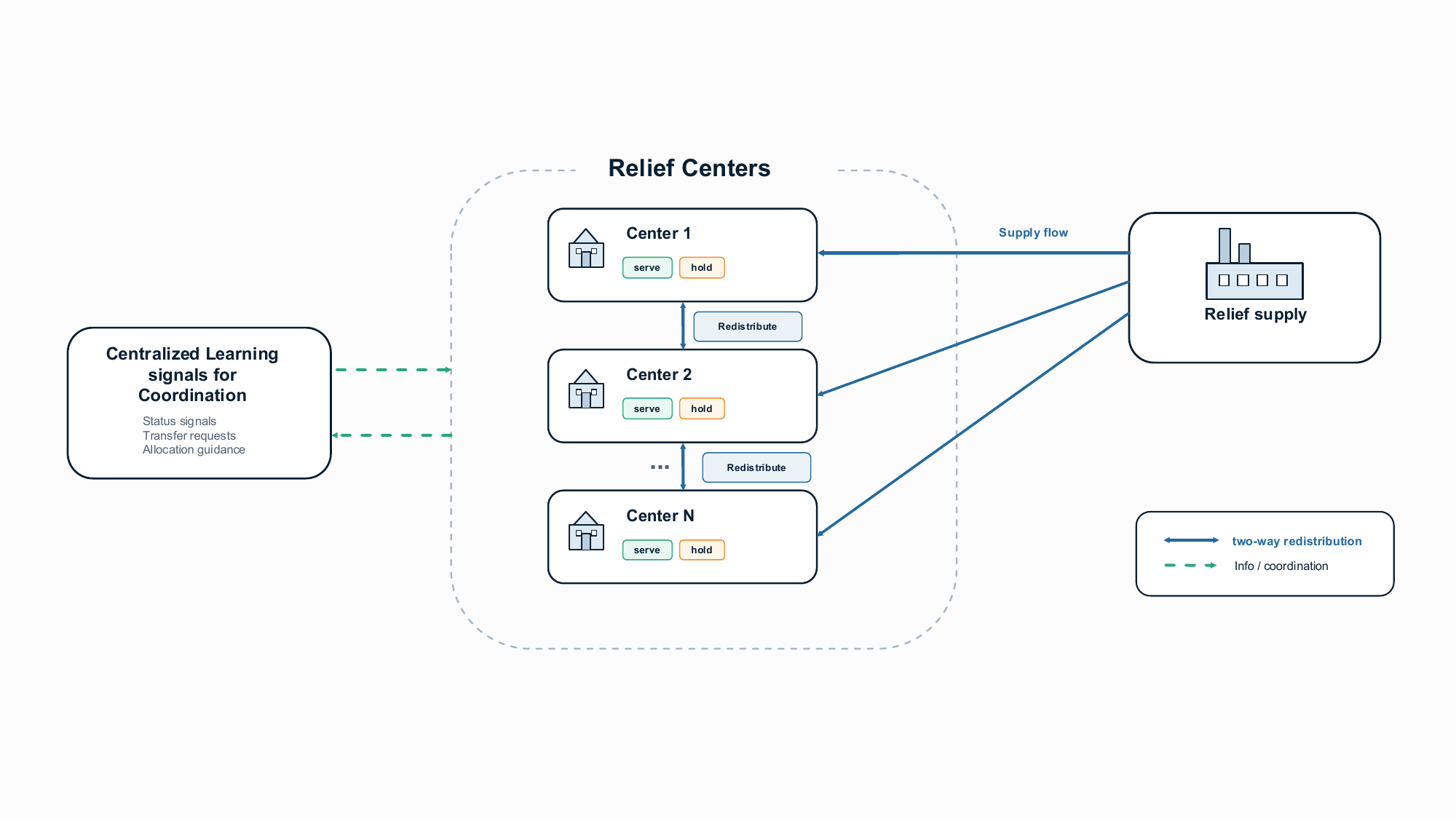}
    \caption{A standard relief supply network with regional relief centers that serve local demand, hold inventory in reserve, and redistribute to other centers.}
    \label{fig:ref_nt}
\end{figure}

We present a relief network problem comprising $N$ regional distribution centers connected by a complete directed transfer graph, such that each center can redistribute to any of the other $N-1$ centers (as shown in Figure \ref{fig:ref_nt}). 
The centers are autonomous, and redistribution decisions are made simultaneously. Each horizon of an emergency event comprises $T$ decision periods (days) that correspond to an acute response to a sudden-onset event. Within each period, exogenous perishable relief supply (lbs) as well as any redistribution whose stochastic travel time, $\tau_{ijt}$, ends that day, arrives at a center. If the total inventory then exceeds capacity, the excess is removed as overflow, oldest stock first. The center then uses its entire available inventory
for local service, a holding reserve, and five specific redistributions to other centers. Finally, held inventory that reaches the end of its shelf life is discarded, and the remaining inventory ages by one day. Table \ref{tab:notation} presents all important notations used in the problem formulation.
\begin{table}[H]
\centering
\caption{Notation used for the relief redistribution model.}
\label{tab:notation}
\footnotesize
\renewcommand{\arraystretch}{1.08}
\setlength{\tabcolsep}{4pt}
\begin{tabularx}{\textwidth}{@{}>{\raggedright\arraybackslash}p{2.3cm}X@{}}
\toprule
\multicolumn{2}{@{}l}{\textbf{Sets and Indices}} \\
\midrule
$\mathcal{N}$
& Set of relief distribution centers; $i,j\in\mathcal{N}$  \\
$T$
& Number of decision periods in the horizon; $t=1,\ldots,T$ [days] \\
$L$
& Shelf life of relief inventory; $\ell=0,\ldots,L-1$ [days] \\
\midrule
\multicolumn{2}{@{}l}{\textbf{Parameters}} \\
\midrule
$D_{it}$
& Demand at center $i$ in period $t$ [lbs] \\
$S_{it}$
& Exogenous perishable supply received by center $i$ in period $t$ [lbs] \\
$K_i$
& Storage capacity of center $i$ [lbs] \\
$\bar{X}_i$
& Daily outbound redistribution capacity of center $i$ [lbs/day] \\
$\tau_{ijt}$
& Travel time for a shipment from center $i$ to center $j$ dispatched in period $t$ [days] \\
$\rho_{ijt}$
& In-transit loss fraction for a shipment from center $i$ to center $j$ dispatched in period $t$  \\
$\zeta_{it}^{\mathrm{trans}}$
& Unobserved transportation disruption at center $i$ in period $t$ \\
$\bar\zeta_{ijt}^{\mathrm{trans}}$
& Link disruption for a shipment from center $i$ to center $j$ in period $t$ [fraction] \\
$\gamma$
& Discount factor \\
\midrule
\multicolumn{2}{@{}l}{\textbf{Variables and Performance Measures}} \\
\midrule
$I_{it}^{(\ell)}$
& Inventory of age $\ell$ at center $i$ at the start of period $t$; $I_{it}^{(0)}=A_{it}$ [lbs] \\
$V_{it}^{(\ell)}$
& Inventory of age $\ell$ at center $i$ in period $t$ after overflow removal [lbs] \\
$H_{it}^{(\ell)}$
& Inventory of age $\ell$ held by center $i$ at the end of period $t$, after service and dispatch [lbs] \\
$A_{it}$
& Total arrivals at center $i$ in period $t$, including exogenous supply and delivered transfers [lbs] \\
$V_{it}$
& Available inventory at center $i$ in period $t$ after arrivals and overflow removal [lbs] \\
$y_{it}$
& Demand served locally at center $i$ in period $t$ [lbs] \\
$u_{it}$
& Unmet demand at center $i$ in period $t$ [lbs] \\
$x_{ijt}$
& Actual dispatched redistribution quantity from center $i$ to center $j$ in period $t$ [lbs] \\
$e_{it}$
& Quantity of expired inventory at center $i$ in period $t$ [lbs] \\
$q_{it}$
& Service ratio of center $i$ in period $t$  \\
$Q_t$
& Network-wide service ratio in period $t$  \\
$E_t$
& Service gap across centers in period $t$  \\
$q_t^{\min}$
& Worst-center service ratio in period $t$  \\
\midrule
\multicolumn{2}{@{}l}{\textbf{MARL Notation}} \\
\midrule
$o_{it}$
& Local observation of center $i$ in period $t$  \\
$a_{it}$
& Continuous allocation action of center $i$ in period $t$  \\
$a_{it}^{\mathrm{loc}}$
& Action share allocated to local service at center $i$  \\
$a_{it}^{\mathrm{hold}}$
& Action share allocated to holding inventory at center $i$  \\
$a_{it}^{j}$
& Action share allocated to intended redistribution from center $i$ to center $j$ \\
$\mathcal{U}_i$
& Simplex action space for center $i$  \\
$b_{it}$
& Recurrent belief state of center $i$ in period $t$ \\
$s_t^{C}$
& Centralized operational state available to the critic in period $t$ \\
$R_t$
& Shared network reward in period $t$ \\
\bottomrule
\end{tabularx}
\end{table}

\subsection{Cooperative Redistribution Model}
\label{sec:method-decpomdp}
We formulate the redistribution problem as a cooperative Markov game, modeled as a decentralized partially observable Markov decision process (Dec-POMDP) \citep{bernstein2002,oliehoek2016}, as specified by the tuple (\ref{eq:decpomdp}).
\begin{equation}
\big\langle\N,\mathcal{S},\{\mathcal{O}_i\}_{i\in\N},
\{\mathcal{U}_i\}_{i\in\N},\mathcal{P},R,\gamma\big\rangle.
\label{eq:decpomdp}
\end{equation}
Here, $\mathcal{N}$ is the set of relief distribution centers (agents) and $\mathcal{S}$ is the
environment state space. $\mathcal{O}_i$ and $\mathcal{U}_i$ are the local observation and action spaces of center $i$, respectively. $\mathcal{P}$ represents transition kernel and $R$ is the shared network reward followed by discount factor $\gamma$. In this problem, the environment state $s_t\in\mathcal{S}$ is not directly
observed by the agents, and they act on local observations
$o_{it}\in\mathcal{O}_i$.

\subsubsection{Transitions}
\label{sec:method-dynamics}

Transitions in the Dec-POMDP are driven by inventory, demand, supply, and
transportation dynamics. We track on-hand inventory by age
$\ell\in\{0,\dots,L-1\}$, where $I_{it}^{(\ell)}$ for $\ell\geq 1$ is the stock carried into day $t$ that has been held for $\ell$ days. All withdrawals from inventory, namely overflow removal, local service and outbound dispatch, take the oldest stock first
\citep{demoor2022}. At the start of day $t$, arrivals at center $i$ include
exogenous supply $S_{it}$ and earlier transfers that arrive on that day, and they enter the youngest age class, $I_{it}^{(0)}=A_{it}$.
\begin{align}
A_{it}
&=S_{it}+\sum_{j\neq i}\sum_{k<t}
(1-\rho_{jik})x_{jik}\mathbbm{1}\{k+\tau_{jik}=t\},
\label{eq:arrivals}\\
\widetilde V_{it}
&=\sum_{\ell=0}^{L-1}I_{it}^{(\ell)}
=A_{it}+\sum_{\ell=1}^{L-1}I_{it}^{(\ell)}.
\label{eq:pre-cap-available}
\end{align}
Here, $\tau_{jik}$ and $\rho_{jik}$ in Eq. \eqref{eq:arrivals} are the realized travel time and loss
fraction for the transfer dispatched from center $j$ to center $i$ on day $k$.
The indicator includes the shipment only on the day of arrival. For a total withdrawal $W$ from an age profile $I^{(\cdot)}$, the oldest-first rule removes from age class $\ell$ the quantity
\begin{equation}
\omega^{(\ell)}\big(W;I^{(\cdot)}\big)
=\min\Big\{I^{(\ell)},\;\max\Big\{0,\;W-\sum_{m=\ell+1}^{L-1}I^{(m)}\Big\}\Big\},
\qquad \ell=0,\ldots,L-1.
\label{eq:fifo}
\end{equation}
If post-arrival
inventory in Eq. \eqref{eq:pre-cap-available} exceeds storage capacity $K_i$, overflow is removed before action, oldest stock first, as shown in Eq. \eqref{eq:available}.
\begin{equation}
\begin{gathered}
O_{it}=\max\{\widetilde V_{it}-K_i,0\},
\qquad
V_{it}^{(\ell)}=I_{it}^{(\ell)}-\omega^{(\ell)}\big(O_{it};I_{it}^{(\cdot)}\big),\\
V_{it}=\sum_{\ell=0}^{L-1}V_{it}^{(\ell)}=\widetilde V_{it}-O_{it}.
\end{gathered}
\label{eq:available}
\end{equation}
Thus, $V_{it}$ is the post-arrival, post-overflow inventory available for
allocation by the policy, and $V_{it}^{(\ell)}$ is its age profile.

\subsubsection{MARL Action}
\label{sec:method-action}

For each relief center $i$, the action is a continuous simplex
$a_{it}\in\mathcal{U}_i$ that represents the fractional allocation of available inventory $V_{it}$ across
local service, holding, and intended redistribution to the other centers, as shown in Eq.~\eqref{eq:action}. All action components are nonnegative, so $a_{it}^{k}\geq 0$ for all $k$, and
$a_{it}^{\mathrm{loc}}+a_{it}^{\mathrm{hold}}+\sum_{j\neq i}a_{it}^{j}=1$.

\begin{equation}
a_{it}=
\left(
a_{it}^{\mathrm{loc}},
a_{it}^{\mathrm{hold}},
\{a_{it}^{j}\}_{j\neq i}
\right)
\label{eq:action}
\end{equation}

Local service is capped by realized demand, as shown in Eq.~\eqref{eq:service}.
\begin{equation}
y_{it}=\min\{D_{it},a_{it}^{\mathrm{loc}}V_{it}\},
\label{eq:service}
\end{equation}
Unmet demand is shown in Eq.~\eqref{eq:unmet}.
\begin{equation}
u_{it}=D_{it}-y_{it}.
\label{eq:unmet}
\end{equation}
For each partner $j\neq i$, the action implies an intended outbound
redistribution, represented by the nominal quantity
$\widetilde{x}_{ijt}=a_{it}^{j}V_{it}$. The total nominal outbound
redistribution from center $i$ is shown in Eq.~\eqref{eq:nom_sup}.
\begin{equation}
\sigma_{it}=\sum_{j\neq i}\widetilde{x}_{ijt}.
\label{eq:nom_sup}
\end{equation}
Nominal redistribution quantities are proportionally scaled only if they exceed
the daily outbound redistribution capacity $\bar X_i$:
\begin{equation}
c_{it}=\min\left\{1,\frac{\bar X_i}{\sigma_{it}+\varepsilon}\right\},
\qquad
x_{ijt}=c_{it}\widetilde{x}_{ijt}.
\label{eq:transfer-scaling}
\end{equation}
The realized held inventory is the residual after local service and dispatch:
\begin{equation}
H_{it}=V_{it}-y_{it}-\sum_{j\neq i}x_{ijt}.
\label{eq:held}
\end{equation}

\subsubsection{Inventory Aging and Expiration}
\label{sec:method-aging}
Local service and outbound dispatch are drawn from the age profile $V_{it}^{(\ell)}$ by the oldest-first rule in Eq. \eqref{eq:fifo}. Here, local service is withdrawn first, and the dispatched quantities are then taken from the oldest remaining stock. Because successive oldest-first withdrawals remove the same stock as a single oldest-first withdrawal of their sum, the age profile of the held inventory is defined in Eq. \eqref{eq:held}.
\begin{equation}
H_{it}^{(\ell)}=V_{it}^{(\ell)}-\omega^{(\ell)}\Big(y_{it}+\sum_{j\neq i}x_{ijt};\,V_{it}^{(\cdot)}\Big),
\qquad
\sum_{\ell=0}^{L-1}H_{it}^{(\ell)}=H_{it}.
\label{eq:held-age}
\end{equation}
At the end of day $t$, held stock that has reached the end of its shelf life expires, and the remaining stock ages by one day as defined in Eq. \ref{eq:aging}.
\begin{equation}
e_{it}=H_{it}^{(L-1)},
\qquad
I_{i,t+1}^{(\ell+1)}=H_{it}^{(\ell)},\quad \ell=0,\ldots,L-2.
\label{eq:aging}
\end{equation}
A unit that arrives on day $t$ can therefore be used on days $t,\ldots,t+L-1$ and is discarded at the end of day $t+L-1$ if it is still held. 
\subsubsection{Local Observations}
\label{sec:method-obs}
Each agent ${i\in\N}$ observes only local states and limited, delayed information about the other agents. The local observation vector for agent $i$ in period $t$ is presented in Eq.~\eqref{eq:observation}.
\begin{equation}
\begin{aligned}
o_{it}=\Big[\;&
\tfrac{V_{it}^{(0:L-1)}}{K_i},\;
\tfrac{V_{it}}{K_i},\;
\tfrac{D_{it}}{\bar D_i},\;
\tfrac{S_{it}}{\bar S_i},\;
\tfrac{u_{i,t-1}}{\bar D_i},\;
\tfrac{P_{it}^{\mathrm{pipe}}}{\bar D_i},\;
q_{i,t-1},\\[1pt]
&\tfrac{\sum_j x_{ij,t-1}}{\bar D_i},\;
\tfrac{\sum_j x_{ji,t-1}}{\bar D_i},\;
\tfrac{t}{T},\;
\mathbf q_{t-1},\;
a_{i,t-1}\;\Big],
\end{aligned}
\label{eq:observation}
\end{equation}

$V_{it}^{(0:L-1)}/K_i$ presents the post-arrival, post-overflow inventory age profile up to the shelf-life cycle, and $V_{it}/K_i$ is the total inventory on hand, both normalized by the center's storage capacity $K_i$. $D_{it}/\bar D_i$ and $S_{it}/\bar S_i$ present the current demand and current supply relative to the center's historical average levels, $\bar D_i$ and $\bar S_i$. The center also records $u_{i,t-1}/\bar D_i$, the unmet demand from the previous day of its demand. Next, $P_{it}^{\mathrm{pipe}}/\bar D_i$ is the quantity already dispatched toward the center but not yet arrived. Here, the inbound pipeline $P_{it}^{\mathrm{pipe}}$ denotes the total quantity
previously dispatched to center $i$ by other centers that remain in transit
when center $i$ forms its period-$t$ observation. It is presented in Eq. \eqref{eq:pipeline}.
\begin{equation}
P_{it}^{\mathrm{pipe}}
=
\sum_{j\neq i}\sum_{k<t}
x_{jik}\,
\mathbbm{1}\!\left\{k+\tau_{jik}>t\right\}.
\label{eq:pipeline}
\end{equation}
The indicator in Eq. \ref{eq:pipeline} is evaluated using observed delivery confirmations, without revealing future shipment arrival times to the receiving center. The inbound pipeline represents the total quantity dispatched toward center $i$ yet to be delivered. Dispatch notices provide the nominal shipment quantities, while delivery confirmations identify shipments that have arrived. The receiving center does not observe the realized transportation delays or in-transit losses before delivery. Following this, $\sum_j x_{ij,t-1}/\bar D_i$ is the quantity the center dispatched to its partners on the previous day, and $\sum_j x_{ji,t-1}/\bar D_i$ is the quantity its partners dispatched toward it on the previous day, as announced in their dispatch notices. Because transfers take one to four days, the latter is not the quantity received on the previous day; received transfers enter the observation through the youngest age class $V_{it}^{(0)}$. Finally, $t/T$ is the current day within the planning horizon of length $T$, and $a_{i,t-1}$ is the center's
previous action. 
In the observation, the only network-wide information available to an agent is $\mathbf q_{t-1}=(q_{1,t-1},\ldots,q_{N,t-1})$, the previous day's
service ratios across the network (defined in Eq. \eqref{eq:center-service}). Agents do not observe partner inventories, demands, supplies, or current actions. Additionally, the current observation alone does not characterize the evolving system state, motivating the recurrent policy described in Section~\ref{sec:method-learning}.

\subsubsection{Shared Reward}
\label{sec:method-reward}

For the reward, we primarily focus on service and equity within the network. Service in a region covered by a center is measured as the fraction of demand met by the center, as shown in Eq. \eqref{eq:center-service}. The aggregated network service is therefore given by Eq. \eqref{eq:network-service}.
\begin{equation}
q_{it}=\frac{y_{it}}{D_{it}+\varepsilon}.
\label{eq:center-service}
\end{equation}

\begin{equation}
Q_t=\frac{\sum_i y_{it}}{\sum_i D_{it}+\varepsilon}.
\label{eq:network-service}
\end{equation}
Equity is measured by the service gap, defined as the mean absolute deviation
of center-level service ratios from their network mean, as shown in Eq. \eqref{eq:equity}.
\begin{equation}
E_t=\frac{1}{N}\sum_i\left|q_{it}-\bar q_t\right|,\qquad
\bar q_t=\frac{1}{N}\sum_i q_{it}.
\label{eq:equity}
\end{equation}
We also measure worst-center service with $q_t^{\min}=\min_i q_{it}$. Finally, we develop a shared reward for all agents shown in Eq. \eqref{eq:reward}.
\begin{equation}
R_t=
w_{1}q_t^{\min}
-w_{2}E_t
-w_{3}\max\{0,\phi-Q_t\}.
\label{eq:reward}
\end{equation}
The first term ensures improvement in the least-served center, while the second penalizes disparities in service across centers. We use the third term to penalize the network service only when $Q_t$ falls below the threshold $\phi$. The third term protects the agents from serving demands rather than serving all centers equally poorly to achieve equity. We use $w_{1},w_{2},w_{3}$ as the weights for the reward terms. For this study, we use $w_{3} \geq w_{2} \geq w_{1}$. The cooperative objective is presented in Eq. \eqref{eq:objective}.
\begin{equation}
J(\boldsymbol\pi)=
\E_{\boldsymbol\pi}
\left[
\sum_{t=0}^{T-1}\gamma^{t}R_t
\right].
\label{eq:objective}
\end{equation}

\subsection{Learning Framework}
\label{sec:method-learning}
 Since we have independent agents operating in a decentralized network, we use MAPPO \citep{YuMAPPO2022}, which supports the cooperative decision-making that this study aims to achieve. MAPPO provides CTDE \citep{kopic2024collaborative}, using a shared critic and decentralized actors. During training, the critic observes the centralized operational state in Eq. \eqref{eq:critic-state}, which includes local observations of all agents, the previous transfer matrix, and the origin-destination pipeline redistribution, where $P_{ijt}^{\mathrm{pipe}}$ is the in-transit quantity from center $i$ to center $j$, so that $\sum_{i\neq j}P_{ijt}^{\mathrm{pipe}}=P_{jt}^{\mathrm{pipe}}$. At execution time, the critic is discarded, and each center acts on its own observations.
\begin{equation}
s_t^{C}=\Big[o_{1t},\dots,o_{Nt},\;
\{x_{ij,t-1}\}_{i,j\in\N},\;
\{P_{ijt}^{\mathrm{pipe}}\}_{i,j\in\N}\Big].
\label{eq:critic-state}
\end{equation}
The critic is therefore a centralized operational-information critic: it pools the operational records that the network generates, but it does not observe the simulator latents, namely the hidden trajectory family, the onset day, the episode severity, the center roles, or the arrival days of in-transit shipments. Hence $s_t^{C}$ is not the complete underlying Markov state $s_t$.

\subsubsection{Recurrent Actor and Policy Update}
Each center (agent) maintains a memory state that is updated at each period. This memory state summarizes the history of observations seen by the agents \citep{cho2014}. We define this by deriving a gated recurrent unit (GRU) \citep{chung2014empirical} as shown in Eq.~\eqref{eq:gru}. For each agent, it summarizes the current observation and the previous memory update that provides the agent with the trajectory dynamics without any direct signal in the observation.
\begin{equation}
b_{it}=\mathrm{GRU}_{\theta_i}(o_{it},b_{i,t-1})
\label{eq:gru}
\end{equation}
The action distribution over the simplex in Eq.~\eqref{eq:action} shows how agents combine the memory state and observation in their policy.
\begin{equation}
\pi_{\theta_i}\!\left(a_{it}\mid o_{it},b_{i,t-1}\right)
=\mathrm{Dirichlet}\!\left[\mathrm{softplus}(W_i b_{it}+c_i)+\alpha_0\right].
\label{eq:dirichlet}
\end{equation}
This Dirichlet distribution \citep{ng2011dirichlet} allows distributions over the simplex that, by construction, have shares that sum to one, thus satisfying the constraint in Eq. \eqref{eq:action}. The offset $\alpha_0>0$ prevents concentration parameters from becoming zero. 

Actors are trained with PPO \citep{Schulman2017}, which updates the policy
while not changing drastically from the previous update. For an action
$a_{it}$ taken by agent $i$ at time $t$, the probability ratio is defined in Eq. \eqref{eq:ratio}.
\begin{equation}
r_{it}(\theta_i)=
\frac{
\pi_{\theta_i}(a_{it}\mid o_{it},b_{i,t-1})
}{
\pi_{\theta_i^{\mathrm{old}}}(a_{it}\mid o_{it},b_{i,t-1}^{\mathrm{old}})
}.
\label{eq:ratio}
\end{equation}
Here, $\pi_{\theta_i}$ is the current policy and
$\pi_{\theta_i^{\mathrm{old}}}$ is the policy before the current update. The clipped PPO objective is shown in Eq. \eqref{eq:ppo}, where the advantage estimate $\widehat A_t$ is used to determine whether an action
should become more or less likely.
\begin{equation}
L_i^{\mathrm{clip}}(\theta_i)
=
\mathbb{E}_t\left[
\min\left\{
r_{it}(\theta_i)\widehat A_t,
\mathrm{clip}\left(
r_{it}(\theta_i),1-\epsilon,1+\epsilon
\right)\widehat A_t
\right\}
\right].
\label{eq:ppo}
\end{equation}
The actor minimizes the clipped negative objective with entropy regularization, which promotes exploration, in Eq. \eqref{eq:actor-loss}.
\begin{equation}
\mathcal{L}_i^{\pi}(\theta_i)
=
-L_i^{\mathrm{clip}}(\theta_i)
-\eta\,
\mathbb{E}_t\left[
\mathcal{H}\left(
\pi_{\theta_i}(\cdot\mid o_{it},b_{i,t-1})
\right)
\right],
\label{eq:actor-loss}
\end{equation}
Here, $\epsilon>0$ controls the clipping range and $\eta$ is the entropy
coefficient.

$\widehat A_t$ is computed using generalized advantage estimation (GAE)
\citep{schulman2016}. The centralized critic $V_\phi(s_t^C)$ estimates the
expected future return from the centralized training state $s_t^C$. After observing
reward $R_t$ and the next state, the one-step temporal-difference error is defined in Eq. \eqref{eq:td-error}.
\begin{equation}
\delta_t
=
R_t+\gamma V_\phi(s_{t+1}^{C})-V_\phi(s_t^{C}),
\label{eq:td-error}
\end{equation}
where $\gamma\in[0,1]$ is the discount factor. GAE combines these prediction errors over subsequent periods using Eq. \eqref{eq:gae}.
\begin{equation}
\widehat A_t
=
\sum_{l=0}^{T-t-1}
(\gamma\lambda)^l\delta_{t+l},
\label{eq:gae}
\end{equation}
where $\lambda\in[0,1]$ controls how strongly future prediction errors
contribute to the current advantage estimate.

The centralized critic is trained to estimate the return target
$\widehat V_t$ computed from the sampled trajectory. Its parameters $\phi$ are learned by minimizing the critic value loss in Eq. \eqref{eq:value-loss}.
\begin{equation}
\mathcal{L}^{V}(\phi)
=
c_V\,
\mathbb{E}_t\left[
\left(
V_\phi(s_t^{C})-\widehat V_t
\right)^2
\right],
\label{eq:value-loss}
\end{equation}
Here, $c_V$ is the value-loss coefficient. The learning framework is summarized in Algorithm~\ref{alg:mappo}.
\begin{algorithm*}[!t]
\caption{Recurrent MAPPO}
\label{alg:mappo}
\small
\begin{algorithmic}[1]
\State Initialize actor parameters $\{\theta_i\}_{i\in\mathcal{N}}$
\State Initialize centralized critic parameters $\phi$
\State Initialize experience buffer $\mathcal{D}\gets\emptyset$
\For{$k=1,\ldots,K$}
 \State $\mathcal{D}\gets\emptyset$
 \For{$e=1,\ldots,N_{\mathrm{ep}}$}
 \State Draw the next realization from $\mathcal{B}_{\mathrm{tr}}$
 \State Reset environment and initialize $b_{i,-1}\gets\mathbf{0}$ for all $i\in\mathcal{N}$
 \For{$t=0,\ldots,T-1$}
 \State Construct observation $o_{it}$ using Eq.~\eqref{eq:observation}
 \State Update recurrent belief
 $b_{it}$ using Eq.~\eqref{eq:gru}
 \State Sample action $a_{it}$ according to Eq.~\eqref{eq:dirichlet}
 \State Execute $a_t$ in the environment 
 \State Observe reward $R_t$ from Eq.~\eqref{eq:reward}
 \State Store
 $(o_{it},a_{it},b_{i,t-1},s_t^{C},R_t)$ in $\mathcal{D}$
 \EndFor
 \If{the complete training-set cycle is finished}
 \State Reshuffle $\mathcal{B}_{\mathrm{tr}}$
 \EndIf
 \EndFor
 \State Estimate advantages $\widehat{A}_t$ using Eq.~\eqref{eq:gae}
 \State Normalize $\widehat{A}_t$ over $\mathcal{D}$
 \For{$j=1,\ldots,N_{\mathrm{epoch}}$}
 \For{each minibatch containing $M$ complete episodes}
 \State Replay recurrent sequences while preserving temporal order
 \State Update actors $\{\theta_i\}$ by ascending the PPO objective
 in Eq.~\eqref{eq:ppo}
 \State Update critic $\phi$ by minimizing the value loss
 in Eq.~\eqref{eq:value-loss}
 \EndFor
 \EndFor
 \If{checkpoint interval is reached}
 \State Evaluate actors on the validation set
 \EndIf
\EndFor
\State \Return trained actors $\{\theta_i\}$
\end{algorithmic}
\end{algorithm*}

\section{Computational Study Design}
\label{sec:exp}
To demonstrate the proposed model's performance, we use two testbeds. First, we develop a synthetic environment with disruptions, generated from hidden trajectory families. After that, we build a data-informed study of Hurricane Ian (2022) across six Southwest Florida counties, based on recorded outages, assistance registrations, and relief facilities operated during the response.
\subsection{Synthetic disruptions}
\label{sec:exp-synthetic}
The network contains $N=6$ relief centers ($i\in\mathcal{N}$) connected by a complete transfer graph, and each episode lasts $T=20$ daily periods of an emergency event. Simulator setup and synthetic parameter values are given in Table \ref{tab:simulator}. In each period $t$, every agent observes a 29-dimensional local observation $o_{it}\in\mathcal{O}_i$ and selects a 7-dimensional simplex action $a_{it}\in\mathcal{U}_i$, consisting of the local-service share $a_{it}^{\mathrm{loc}}$, the holding share $a_{it}^{\mathrm{hold}}$ and the redistribution shares $\{a_{it}^{j}\}_{j\neq i}$ to the other five centers. The MAPPO critic uses the 246-dimensional centralized operational state $s_t^C$ defined in Eq.~\eqref{eq:critic-state}.

Each synthetic episode draws one of five hidden trajectory families (early recovery, delayed peak, progressive worsening, persistent disruption and aftershock), an onset day, an episode severity and a random assignment of two shortage, two surplus and two balanced center roles; the roles determine how strongly the disruption raises demand, reduces supply and delays transport at each center (\ref{app:generator}). The trajectory families and the roles are not provided explicitly to the actors or the critic; they are reflected only indirectly through their effects on the observed demand, supply, service and transfers. A fixed bank of 200 realizations is divided into 160 training, 20 validation and 20 test realizations, with each family represented equally in every split.

\subsection{Hurricane Ian Environment}
\label{sec:exp-ian}
\begin{figure*}[!ht]
\centering
\includegraphics[width=\textwidth]{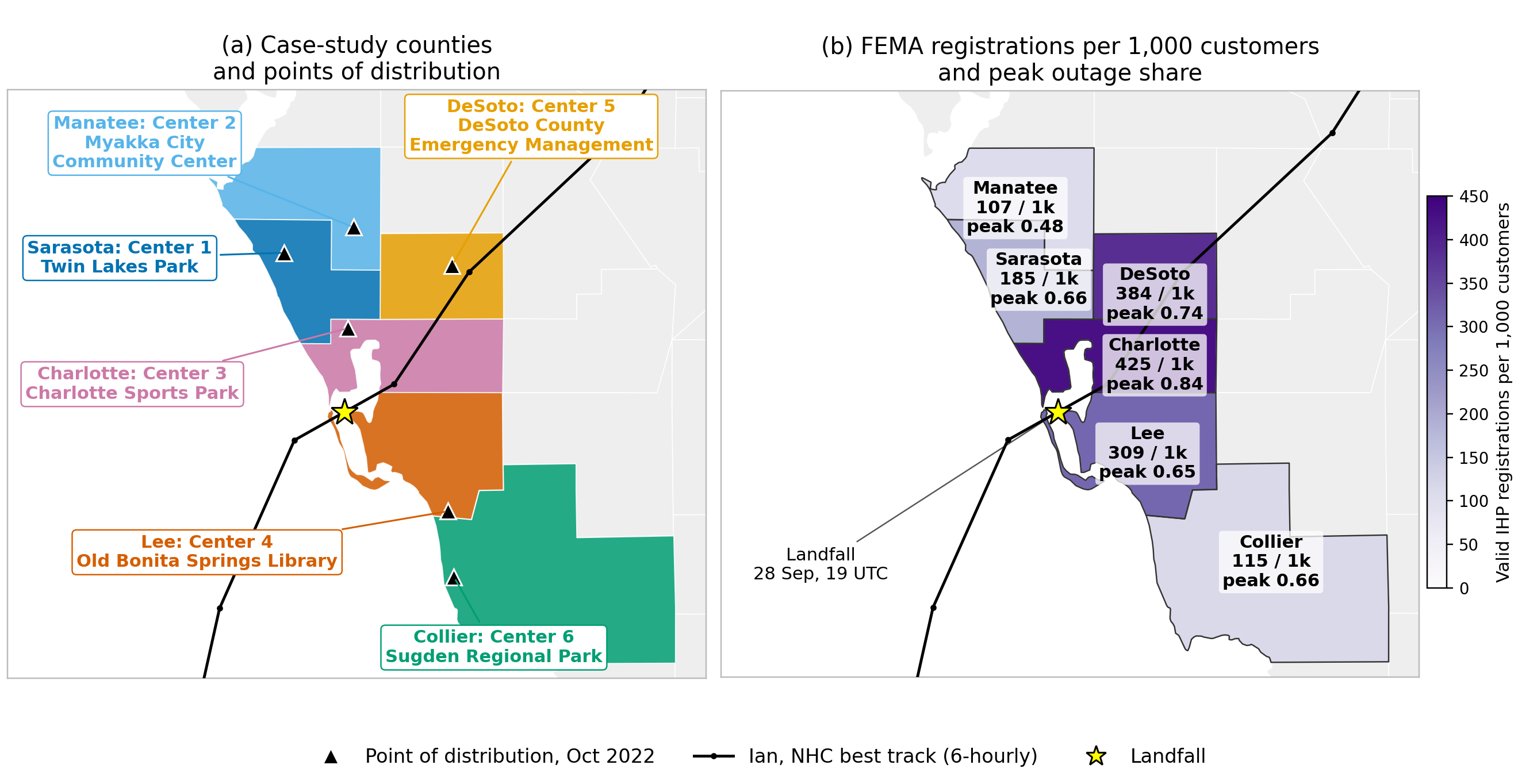}
\caption{Hurricane Ian case study with (a) the six counties, the point of distribution for the county's relief center; the line is the National Hurricane Center best track at six-hour intervals. The star marks landfall on Cayo Costa. (b) FEMA valid Individual and Households Program registrations per 1000 electricity customers and the peak daily share of customers without power (EAGLE-I) in each county.}
\label{fig:ian-map}
\end{figure*}

Hurricane Ian affected Cayo Costa in Lee County on 28 September 2022, and Southwest Florida was hit hardest \citep{Bucci2023}. FEMA declared a major disaster (DR-4673-FL) and operated points of distribution (PODs) that handed out emergency supplies to residents. We use this as our second testbed with an observed disruption. The case study consists of six neighboring counties around the hurricane: Lee, Charlotte, Sarasota, Manatee, DeSoto, and Collier, as shown in Figure \ref{fig:ian-map}.

In each county, we use one distribution point that operated in early October 2022 as the county's relief center. The case-study environment keeps the trained policies unchanged and replaces the synthetic disruption from the synthetic environment in \ref{sec:exp-synthetic}. Each center follows the observed daily outage share for its county from the EAGLE-I county outage records \citep{Brelsford2024}, rather than a hidden trajectory family. Here, electricity outages serve as a proxy for the severity of local disruptions, as power failures can disrupt food storage, retail operations, and access to essential supplies, thereby increasing the need for emergency assistance. Its demand surge, supply reduction, and delayed relief response are scaled by the county's FEMA Individuals and Households Program registrations per 1000 electricity customers \citep{OpenFEMA2026}. Thus, the observed outage trajectories determine the timing of disruptions, while FEMA registrations provide a measure of their relative severity across counties. The baseline supply of each center is not observed for the Florida sites; as in the synthetic environment, it is bootstrapped from the historical daily receipts of a six-branch food bank network (2018--2021), sampling whole weekday-matched dates on the 2022 calendar to preserve the correlation across centers, and the demand and supply noise of Eqs.~\eqref{eq:demand} and \eqref{eq:supply} is applied on top. The horizon covers the 20 days from 26 September to 15 October 2022. \ref{app:ian} details the data, the sites, and the construction of the environment.

The six counties were affected differently, as shown by the peak outages and registrations in Figure \ref{fig:ian-map}(b) and by the daily outage paths in Figure \ref{fig:ian}(a). None of these county-specific paths belong to the trajectory families used in training. The case is used only for evaluation, and the policies trained on the synthetic disruptions are applied to it without retraining.

\subsection{Baselines}
\label{sec:exp-baselines}
We compare recurrent MAPPO with four alternatives that act in the same environment on the same realizations. First, we use recurrent IPPO that shares the actor, action space, reward and optimizer settings of MAPPO but replaces the centralized critic with a local critic for each agent. Second, a Local-only operation that serves local demand from available inventory and never redistributes. Next, Uniform surplus sharing rule serves local demand first and ships the post-service surplus $V_{it}-y_{it}$, up to the outbound capacity $\bar X_i$, in equal shares to the five other centers. Finally, a Need-proportional sharing rule that serves local demand first and ships the same quantity to the partners in proportion to their shortfall in the previous period, $1-q_{j,t-1}$.

\subsection{Metrics}
\label{sec:exp-protocol}
The primary metrics are the network service $Q_t$, the worst-center service $q_t^{\min}$, the service gap $E_t$ and the daily reward $R_t$, each averaged over the $T$ periods of an episode; we also report the quantity redistributed over the episode relative to total demand. MAPPO and IPPO are trained for 30{,}000 episodes on the 160 training realizations. Their hyperparameters are set with grid search and are listed in Table \ref{tab:hyper}. During training, the policies are evaluated every 160 episodes on the 20 validation realizations. The final policies are then evaluated once, with deterministic actions, on the 20 held-out test realizations and on 20 noise realizations of the Hurricane Ian case, which differ in demand, supply, transport delays and losses. All methods are evaluated on the same realizations, so differences in performance are attributable to the decision policies, and pairwise differences are tested with paired two-sided $t$-tests across the common realizations.
\section{Results and Discussion}
\label{sec:results}
We first report the learning dynamics and the performance on the synthetic test bed and then the Hurricane Ian case.

\subsection{Training and validation}
\label{sec:res-learning}
\begin{figure*}[!ht]
\centering
\includegraphics[width=\textwidth]{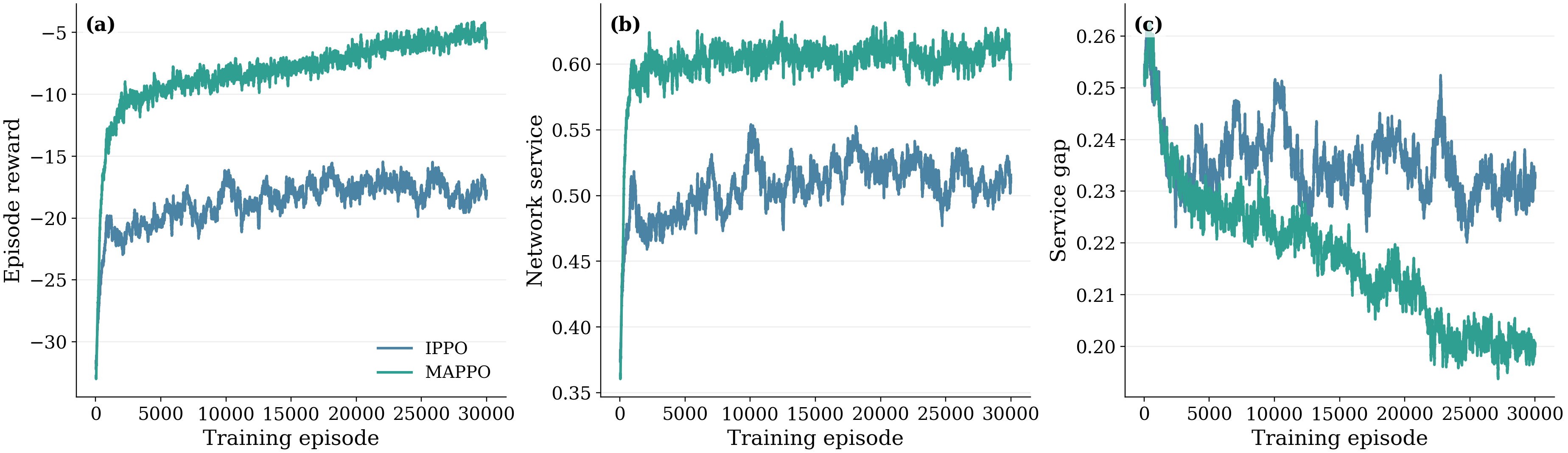}
\caption{On-policy training dynamics presenting (a) episodic reward, (b) mean daily network service, and (c) mean daily service gap. Curves are hundred-episode moving averages from sampled training actions.}
\label{fig:training}
\end{figure*}

Figure \ref{fig:training}(a)-(c) presents on-policy training dynamics over 30000 episodes for both MAPPO and IPPO, reporting episodic return $\sum_{t=0}^{T-1}\gamma^{t}R_t$, mean daily network service $\bar Q_t= \frac{\sum_t Q_t}{T}$, and mean daily service gap $\bar E_t=\frac{\sum_t E_t}{T}$, respectively. According to the results, both MAPPO and IPPO improve over their initial policies, but MAPPO achieves a stronger final policy with a mean of the last 100 episode rewards of 
$-5.68$ compared with $-17.81$ for IPPO, corresponding to a $68.1\%$ smaller reward penalty. MAPPO also achieves a higher network service $\bar Q_t$ of $0.599$, representing a $16.2\%$ relative improvement over IPPO, while reducing the service gap by $14.1\%$, ensuring better equity in the network.
\begin{figure*}[!ht]
\centering
\begin{minipage}[t]{0.49\textwidth}
\centering
\includegraphics[width=\textwidth]{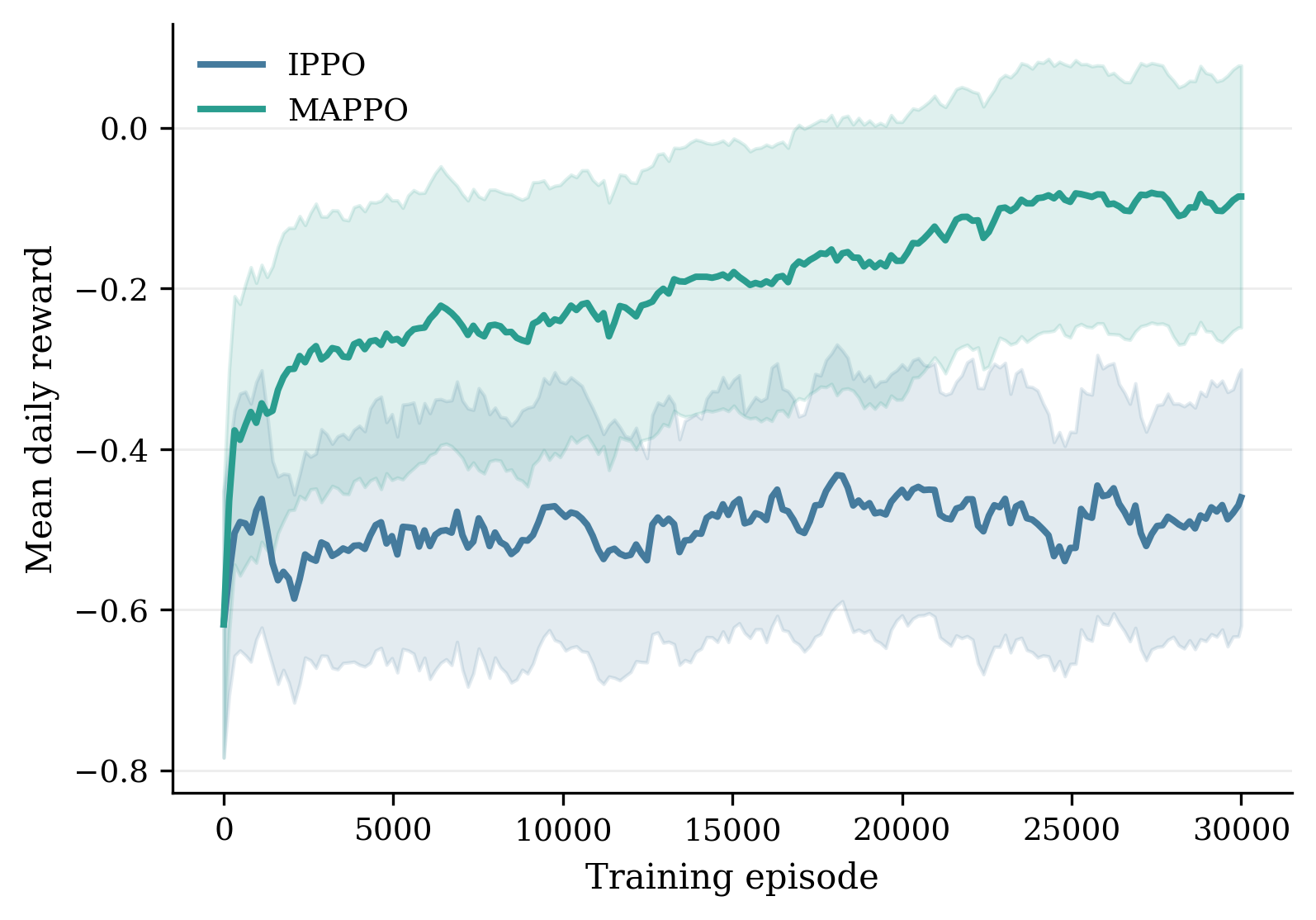}\\[2pt]
{\footnotesize (a) }
\end{minipage}
\hfill
\begin{minipage}[t]{0.49\textwidth}
\centering
\includegraphics[width=\textwidth]{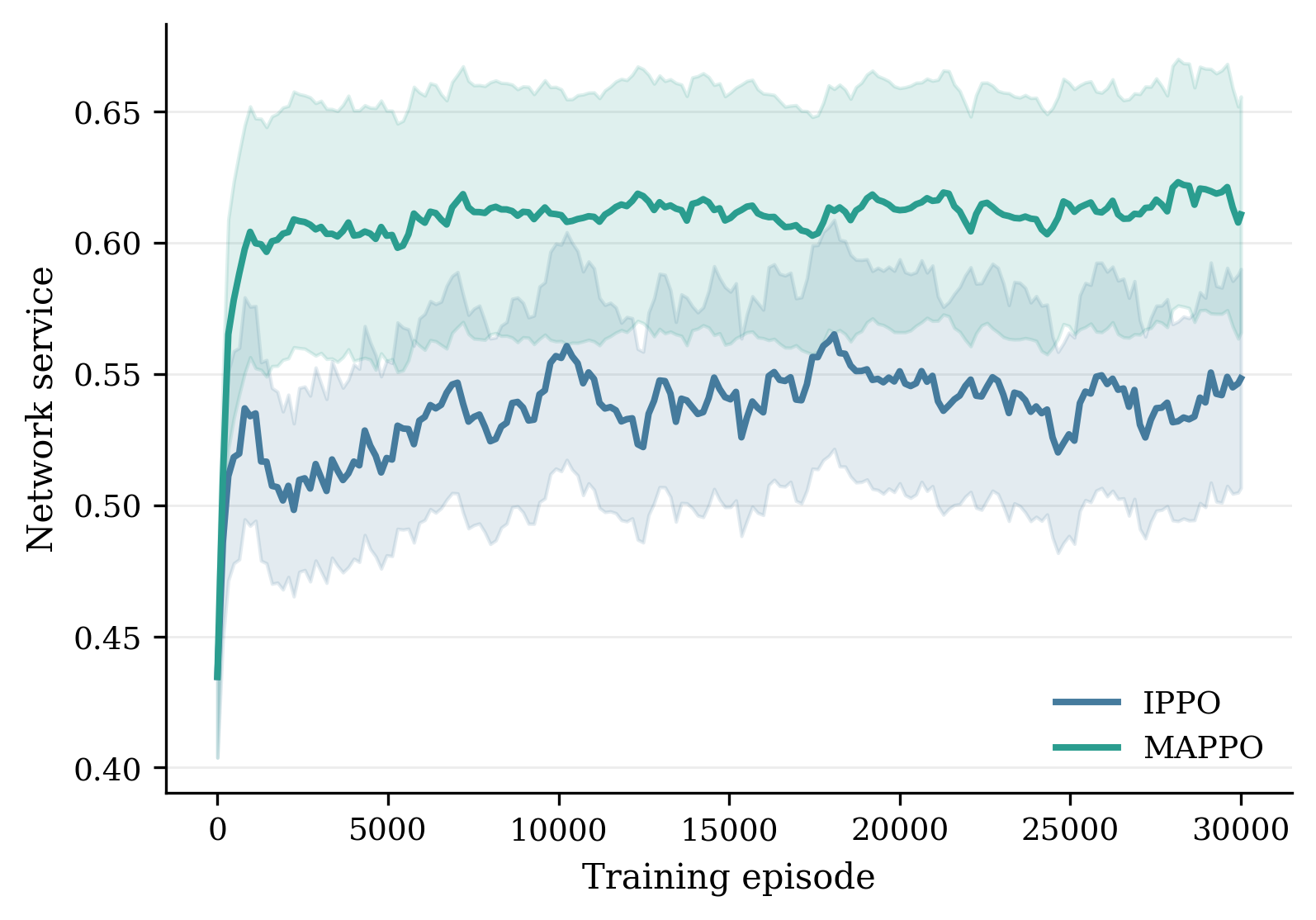}\\[2pt]
{\footnotesize (b) }
\end{minipage}
\begin{minipage}[t]{0.49\textwidth}
\centering
\includegraphics[width=\textwidth]{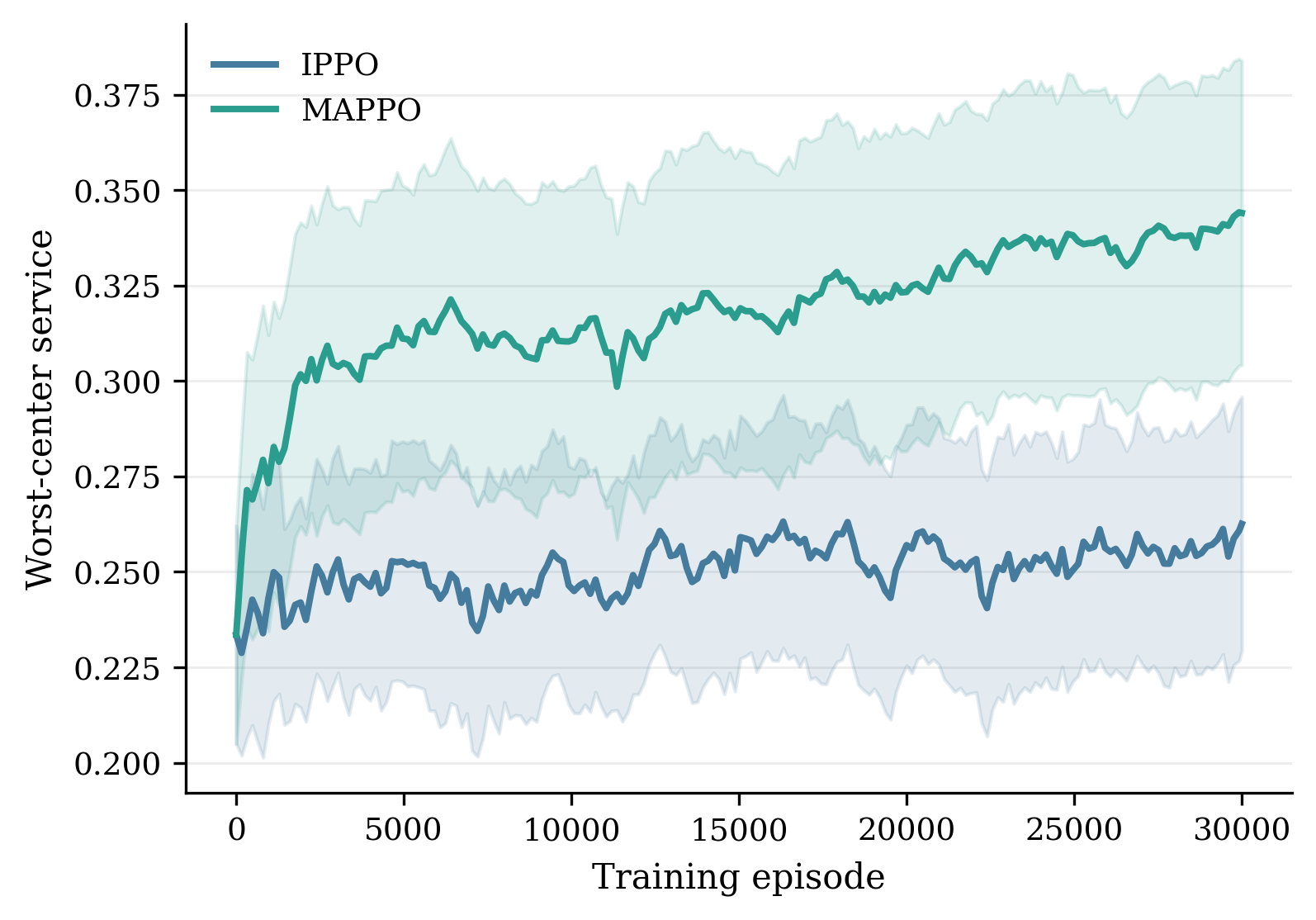}\\[2pt]
{\footnotesize (c) }
\end{minipage}
\hfill
\begin{minipage}[t]{0.49\textwidth}
\centering
\includegraphics[width=\textwidth]{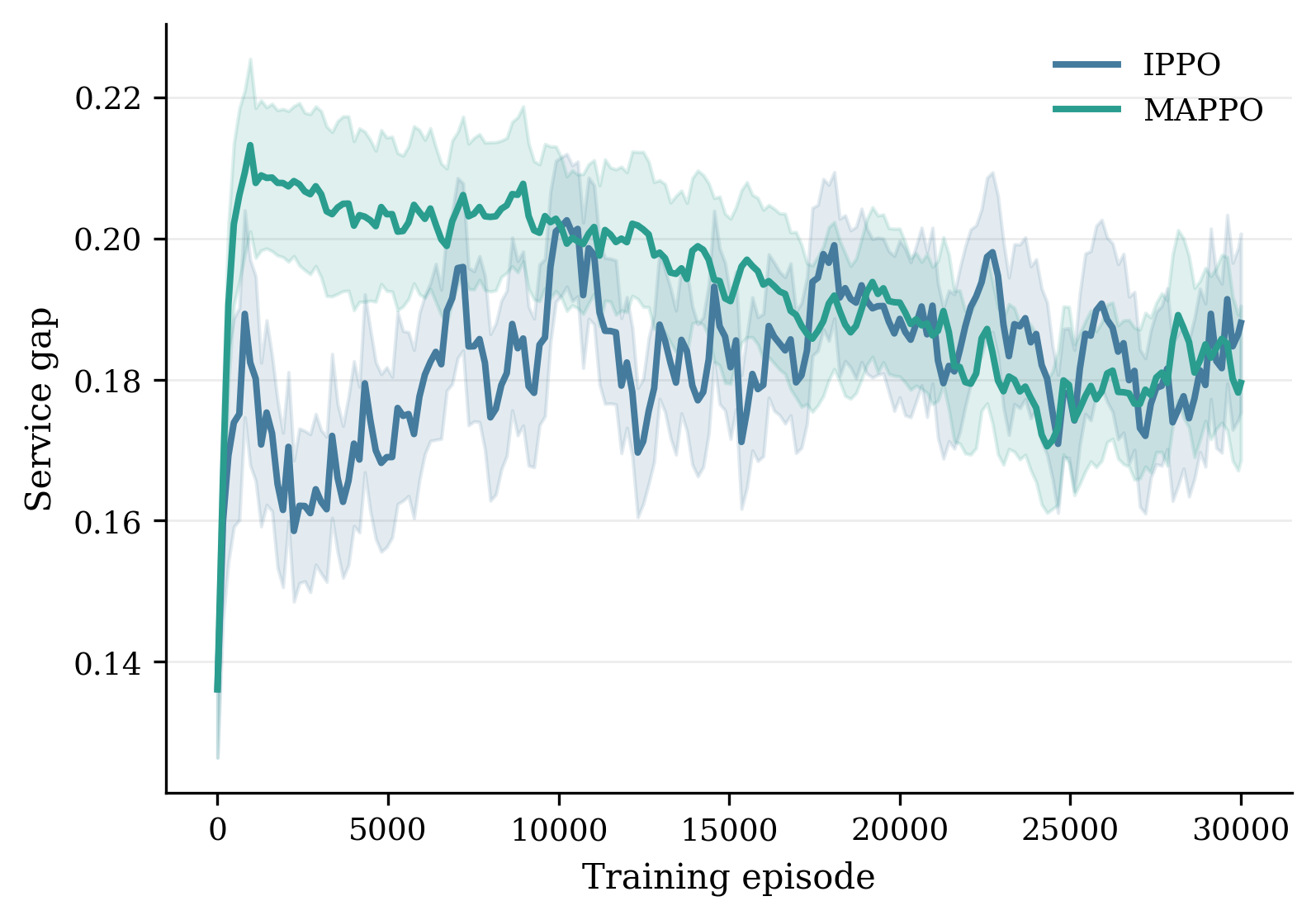}\\[2pt]
{\footnotesize (d) }
\end{minipage}
\caption{Trained policies evaluated every 160 training episodes on
the same 20 validation set. (a) Mean daily
reward, (b) network service, (c) worst-center service, and (d) service gap.}
\label{fig:validation}
\end{figure*}
Figure \ref{fig:validation}(a)-(d) shows the evaluation of the trained policies at 160-episode intervals. Both learners start from the same policy, with a small gap across all metrics, because all centers are served poorly. As learning begins, service
improves unevenly across centers, so the service gap rises as shown in Figure \ref{fig:validation} (d). For MAPPO, the gap
peaks at $0.213$ around episode $960$ while the network service has already reached $0.604$ in Figure \ref{fig:validation}(b). After this early high-gap phase, MAPPO continues to improve service while closing the gap and improving equitable distribution and network service, exceeding IPPO. At the final checkpoint, MAPPO has an $11.4\%$ higher network service, a $31.0\%$ higher worst-center service, and a $4.5\%$ lower service gap than IPPO, demonstrating the capability of cooperative learning to improve network service while maintaining a lower service gap, thereby ensuring equitable distribution across the network.

\subsection{Held-out performance}
\label{sec:res-validation-heldout}
Table \ref{tab:heldout} summarizes the performance on the 20 held-out test realizations. These realizations were not used in training, but they are drawn from the same five trajectory families and the same generator as the training realizations. Compared with local-only, MAPPO reduces the service gap by 33.2\% and improves worst-center service by 50.4\%, both of which are statistically significant. In contrast with uniform surplus and need-proportional sharing, MAPPO reduces the service gap by 16.9\% and 19.6\%, both significant. MAPPO also improves worst-center service by 4.2\% and 10.9\% against the two rules; only the improvement over need-proportional sharing is significant. These equity gains come at the expense of network service. MAPPO serves $0.6276$ of network demand, which is significantly lower than both sharing heuristics and also significantly lower than local-only operation ($0.6686$). However, in this humanitarian relief study, we prioritize improving worst-center service and reducing disparities across communities, at the expense of some overall network service, shaped by the MAPPO reward function.

Compared with IPPO, MAPPO improves network service by 12.0\% and worst-center service by 28.7\%, both statistically significant, while reducing the service gap by 1.9\%. MAPPO also achieves the highest average daily reward among the five methods, with statistically significant improvements over all baselines. Overall, MAPPO achieves the lowest service disparity and highest worst-center service, demonstrating the benefits of centralized training for equitable decentralized redistribution.
\begin{table}[H]
\centering
\caption{Performance on the 20 held-out test realizations. Values represent
mean $\pm$ standard deviation over the 20 realizations. Symbols on the
MAPPO row indicate a statistically significant difference from
IPPO ($\dagger$), local-only ($\ddagger$), uniform surplus sharing ($\S$) and need-proportional sharing ($\P$). The last column is the quantity redistributed over the episode relative to total demand.}
\label{tab:heldout}
\footnotesize
\setlength{\tabcolsep}{3pt}
\renewcommand{\arraystretch}{1.15}
\resizebox{\textwidth}{!}{%
\begin{tabular}{@{}lccccc@{}}
\toprule
Method & Network service & Worst-center service & Service gap & Daily reward & Redistributed\\
       & $\bar Q_t$
       & $\bar q_t^{\min}$
       & $\bar E_t$
       & $ \bar R_t$
       & / demand\\
\midrule
Local-only & $0.6686\pm0.0747$ & $0.2369\pm0.0998$
           & $0.2643\pm0.0415$ & $-0.5801\pm0.3800$ & $0.000$\\
Uniform surplus & $0.6977\pm0.0780$ & $0.3420\pm0.1032$
           & $0.2126\pm0.0359$ & $-0.1926\pm0.3741$ & $0.111\pm0.029$\\
Need-proportional & $\mathbf{0.6989}\pm0.0761$ & $0.3211\pm0.0917$
           & $0.2197\pm0.0320$ & $-0.2620\pm0.3476$ & $0.091\pm0.023$\\
IPPO       & $0.5602\pm0.0569$ & $0.2767\pm0.0646$
           & $0.1801\pm0.0395$ & $-0.3786\pm0.2628$ & $0.215\pm0.012$\\
MAPPO      & $0.6276\pm0.0687$\textsuperscript{$\dagger,\ddagger,\S,\P$}
           & $\mathbf{0.3562}\pm0.0753$\textsuperscript{$\dagger,\ddagger,\P$}
           & $\mathbf{0.1766}\pm0.0302$\textsuperscript{$\ddagger,\S,\P$}
           & $\mathbf{-0.0120}\pm0.2510$\textsuperscript{$\dagger,\ddagger,\S,\P$}
           & $0.207\pm0.010$\\
\bottomrule
\end{tabular}%
}
\end{table}

\subsubsection{Performance across trajectory families}
\label{sec:res-family}
We next examine whether the recurrent policies adapt to unknown dynamics using their observation histories. Table \ref{tab:family} reports performance across the five hidden trajectory families (detailed in \ref{app:generator}), with four held-out realizations per family. MAPPO achieves significantly higher network service than IPPO across all families, with improvements ranging from 10.6\% to 15.3\%. It also improves worst-center service by 19.5\%–41.6\%, with statistically significant differences in four families. Compared with local-only operation, MAPPO significantly improves worst-center service and reduces the service gap in all five families.

MAPPO achieves the highest mean worst-center service, the lowest service gap, and the highest daily reward across four families, i.e., delayed peak, persistent disruption, progressive worsening, and aftershock. However, these advantages over the surplus-sharing heuristics are not consistently statistically significant. In early recovery, uniform surplus sharing achieves a higher worst-center service of 0.4733 compared to 0.4354, while IPPO produces the smallest service gap. Both sharing heuristics achieve significantly higher network service than MAPPO across all families, highlighting the trade-off between overall service and equity. 
Overall, MAPPO generalizes to unseen realizations of the trajectory families encountered during training, even without directly observing the active family. 
\begin{table}[H]
\centering
\caption{Held-out performance by hidden trajectory family (four held-out realizations per family). Values represent mean $\pm$ standard deviation. Symbols on the MAPPO rows indicate a statistically significant difference from IPPO ($\dagger$), local-only ($\ddagger$), uniform surplus sharing ($\S$) and need-proportional sharing ($\P$).}
\label{tab:family}
\footnotesize
\setlength{\tabcolsep}{2.5pt}
\renewcommand{\arraystretch}{1.15}
\resizebox{\textwidth}{!}{%
\begin{tabular}{@{}llcccc@{}}
\toprule
Family & Method & Network service & Worst-center service & Service gap & Daily reward\\
       &        & $(\bar Q_t)$ & $(\bar q_t^{\min})$ & $(\bar E_t)$ & $(\bar R_t)$\\
\midrule
Early recovery
 & Local-only        & $0.7437\pm0.0280$ & $0.3531\pm0.0970$ & $0.2242\pm0.0221$ & $-0.1504\pm0.2770$\\
 & Uniform surplus   & $\mathbf{0.7732}\pm0.0308$ & $\mathbf{0.4733}\pm0.0574$ & $0.1750\pm0.0063$ & $\mathbf{0.2453}\pm0.1803$\\
 & Need-proportional & $0.7709\pm0.0303$ & $0.4308\pm0.0406$ & $0.1884\pm0.0077$ & $0.1084\pm0.1673$\\
 & IPPO              & $0.6126\pm0.0508$ & $0.3645\pm0.0257$ & $\mathbf{0.1473}\pm0.0129$ & $-0.0438\pm0.1437$\\
 & MAPPO             & $0.6896\pm0.0507$\textsuperscript{$\dagger,\S,\P$} & $0.4354\pm0.0623$\textsuperscript{$\ddagger,\S$} & $0.1661\pm0.0360$\textsuperscript{$\ddagger$} & $0.2159\pm0.1936$\textsuperscript{$\ddagger$}\\
\midrule
Delayed peak
 & Local-only        & $0.6537\pm0.1262$ & $0.2324\pm0.1268$ & $0.2770\pm0.0564$ & $-0.6444\pm0.5538$\\
 & Uniform surplus   & $\mathbf{0.6714}\pm0.1400$ & $0.3275\pm0.1475$ & $0.2172\pm0.0515$ & $-0.2988\pm0.6002$\\
 & Need-proportional & $0.6706\pm0.1376$ & $0.2989\pm0.1379$ & $0.2233\pm0.0430$ & $-0.3813\pm0.5732$\\
 & IPPO              & $0.5271\pm0.0860$ & $0.2555\pm0.0677$ & $0.1789\pm0.0418$ & $-0.5134\pm0.3908$\\
 & MAPPO             & $0.6079\pm0.1191$\textsuperscript{$\dagger,\ddagger,\S,\P$} & $\mathbf{0.3617}\pm0.1054$\textsuperscript{$\dagger,\ddagger$} & $\mathbf{0.1709}\pm0.0193$\textsuperscript{$\ddagger$} & $\mathbf{-0.0393}\pm0.4104$\textsuperscript{$\dagger,\ddagger,\P$}\\
\midrule
Persistent
 & Local-only        & $0.6775\pm0.0442$ & $0.2451\pm0.0634$ & $0.2529\pm0.0173$ & $-0.5417\pm0.2580$\\
 & Uniform surplus   & $0.6961\pm0.0699$ & $0.3247\pm0.0836$ & $0.2029\pm0.0150$ & $-0.2403\pm0.3523$\\
 & Need-proportional & $\mathbf{0.6995}\pm0.0667$ & $0.3128\pm0.0651$ & $0.2078\pm0.0166$ & $-0.2842\pm0.3258$\\
 & IPPO              & $0.5612\pm0.0586$ & $0.2750\pm0.0416$ & $0.1697\pm0.0395$ & $-0.3752\pm0.1229$\\
 & MAPPO             & $0.6219\pm0.0644$\textsuperscript{$\dagger,\ddagger,\S,\P$} & $\mathbf{0.3530}\pm0.0447$\textsuperscript{$\dagger,\ddagger$} & $\mathbf{0.1628}\pm0.0182$\textsuperscript{$\ddagger,\S$} & $\mathbf{-0.0235}\pm0.1817$\textsuperscript{$\dagger,\ddagger$}\\
\midrule
Progressive worsening
 & Local-only        & $0.6191\pm0.0663$ & $0.1547\pm0.0496$ & $0.2908\pm0.0388$ & $-0.8563\pm0.2266$\\
 & Uniform surplus   & $0.6647\pm0.0501$ & $0.2680\pm0.0508$ & $0.2431\pm0.0273$ & $-0.4020\pm0.1665$\\
 & Need-proportional & $\mathbf{0.6672}\pm0.0505$ & $0.2632\pm0.0648$ & $0.2466\pm0.0172$ & $-0.4319\pm0.2046$\\
 & IPPO              & $0.5399\pm0.0317$ & $0.2309\pm0.0436$ & $0.1977\pm0.0359$ & $-0.5264\pm0.1137$\\
 & MAPPO             & $0.5985\pm0.0383$\textsuperscript{$\dagger,\S,\P$} & $\mathbf{0.3054}\pm0.0569$\textsuperscript{$\dagger,\ddagger$} & $\mathbf{0.1799}\pm0.0256$\textsuperscript{$\ddagger,\S,\P$} & $\mathbf{-0.1075}\pm0.1819$\textsuperscript{$\dagger,\ddagger,\S,\P$}\\
\midrule
Aftershock
 & Local-only        & $0.6487\pm0.0242$ & $0.1988\pm0.0451$ & $0.2766\pm0.0419$ & $-0.7078\pm0.1894$\\
 & Uniform surplus   & $0.6829\pm0.0182$ & $0.3163\pm0.0307$ & $0.2248\pm0.0320$ & $-0.2671\pm0.0471$\\
 & Need-proportional & $\mathbf{0.6863}\pm0.0173$ & $0.2996\pm0.0477$ & $0.2326\pm0.0361$ & $-0.3207\pm0.1222$\\
 & IPPO              & $0.5604\pm0.0206$ & $0.2578\pm0.0584$ & $0.2069\pm0.0463$ & $-0.4342\pm0.1702$\\
 & MAPPO             & $0.6199\pm0.0255$\textsuperscript{$\dagger,\S,\P$} & $\mathbf{0.3255}\pm0.0505$\textsuperscript{$\dagger,\ddagger$} & $\mathbf{0.2033}\pm0.0415$\textsuperscript{$\ddagger$} & $\mathbf{-0.1055}\pm0.1813$\textsuperscript{$\dagger,\ddagger$}\\
\bottomrule
\end{tabular}%
}
\end{table}

\subsection{Hurricane Ian case study}
\label{sec:res-ian}
Table \ref{tab:ian-results} reports the outcomes of the Hurricane Ian case described in Section \ref{sec:exp-ian}. The case study reveals that the equity advantage of MAPPO is clearer than on the synthetic test. MAPPO achieves the smallest service gap of $0.1257$ versus $0.1766$ for uniform surplus sharing, and attains the highest worst-center service with $0.4535$, all of which are statistically significant. However, uniform surplus sharing achieves higher network service, demonstrating the trade-off between aggregate service and equity. MAPPO also significantly outperforms IPPO across all four performance metrics.

\begin{figure*}[!ht]
\centering
\includegraphics[width=\textwidth]{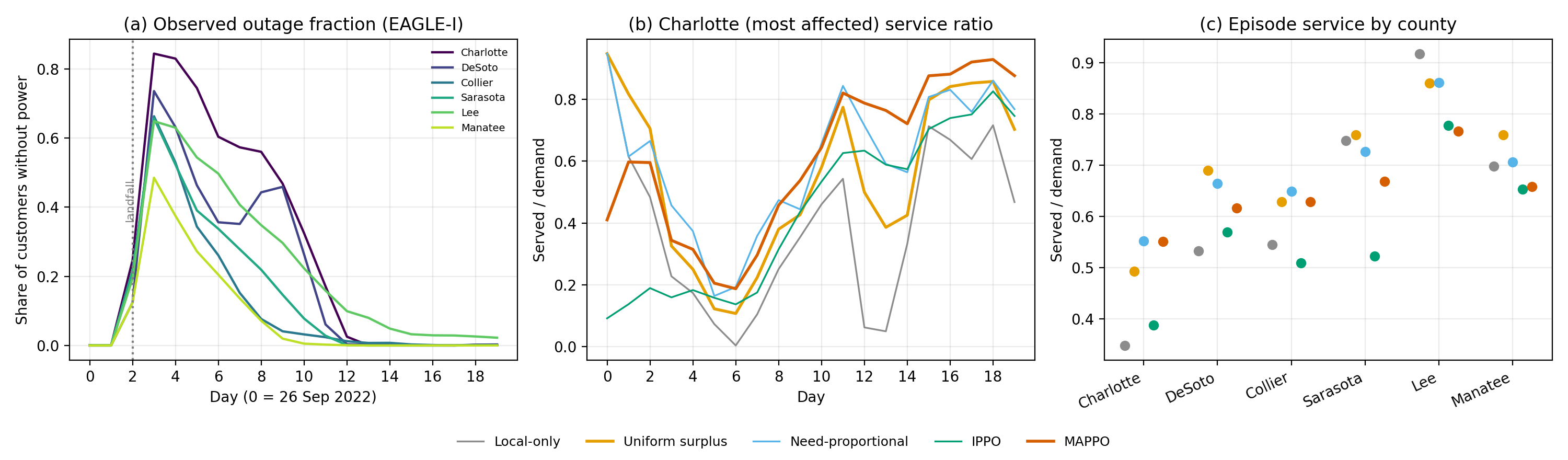}
\caption{Hurricane Ian case study with (a) observed share of electricity customers without power by county; the dotted line marks landfall. (b) Daily service ratio in the most affected county (Charlotte). (c) Service by county over the 20-day horizon. Averages over 20 noise realizations of the data-informed case.}
\label{fig:ian}
\end{figure*}

Moreover, Figure \ref{fig:ian}(c) illustrates the county-level effects of redistribution. MAPPO results in the largest net inflow, equivalent to $31\%$ of local demand, to Charlotte, the most affected county, improving its service from $0.348$ under local-only operation to $0.551$. Figure \ref{fig:ian}(b) presents the service ratio by period for Charlotte. However, this redistribution reduces service in Lee, the surplus county, from $0.918$ to $0.766$. Uniform surplus sharing performs better in DeSoto with $0.690$ against $0.616$ for MAPPO. Overall, MAPPO maintains its equity advantage under observed disruption trajectories not encountered during training, although simple surplus-sharing rules achieve higher aggregate network service.

\begin{table}[H]
\centering
\caption{Hurricane Ian case study: mean $\pm$ standard deviation over 20 noise realizations of the data-informed case. Symbols on the MAPPO row indicate a statistically significant difference from IPPO ($\dagger$), local-only ($\ddagger$), uniform surplus sharing ($\S$) and need-proportional sharing ($\P$).}
\label{tab:ian-results}
\footnotesize
\setlength{\tabcolsep}{3pt}
\renewcommand{\arraystretch}{1.15}
\resizebox{\textwidth}{!}{%
\begin{tabular}{@{}lccccc@{}}
\toprule
Method & Network service & Worst-center service & Service gap & Daily reward & Redistributed\\
       & $\bar Q_t$ & $\bar q_t^{\min}$ & $\bar E_t$ & $\bar R_t$ & / demand\\
\midrule
Local-only        & $0.7255\pm0.0623$ & $0.2756\pm0.0842$ & $0.2453\pm0.0274$ & $-0.3762\pm0.2997$ & $0.000$\\
Uniform surplus   & $\mathbf{0.7489}\pm0.0671$ & $0.4187\pm0.0986$ & $0.1766\pm0.0289$ & $0.0742\pm0.3630$ & $0.127\pm0.019$\\
Need-proportional & $0.7462\pm0.0657$ & $0.3534\pm0.1002$ & $0.2031\pm0.0323$ & $-0.1328\pm0.3582$ & $0.108\pm0.017$\\
IPPO              & $0.6304\pm0.0553$ & $0.3575\pm0.0500$ & $0.1453\pm0.0130$ & $0.0101\pm0.2152$ & $0.222\pm0.003$\\
MAPPO             & $0.6859\pm0.0554$\textsuperscript{$\dagger,\ddagger,\S,\P$}
                  & $\mathbf{0.4535}\pm0.0653$\textsuperscript{$\dagger,\ddagger,\S,\P$}
                  & $\mathbf{0.1257}\pm0.0166$\textsuperscript{$\dagger,\ddagger,\S,\P$}
                  & $\mathbf{0.3750}\pm0.2121$\textsuperscript{$\dagger,\ddagger,\S,\P$}
                  & $0.218\pm0.004$\\
\bottomrule
\end{tabular}%
}
\end{table}

\section{Managerial Insights}
\label{sec:managerial}
From the results discussed in Section \ref{sec:results}, we observe several insights for relief organizations conducting decentralized distribution.

First, uniform surplus sharing can serve as the simplest rule a network can adopt to leverage the benefits of redistribution. Compared with local-only operation, uniform surplus sharing raises the network service from $0.726$ to $0.749$, and the worst-center service from $0.276$ to $0.419$ in the Hurricane Ian case. This rule requires no model or training, and a center only needs to know its own surplus and its outbound capacity. 

On the other hand, the learned policy achieves the most even service, the smallest service gap, and the highest worst-center service on average. However, this comes at the cost of trading off some overall network service. Managers should state whether their priority is the total quantity delivered or the protection of the worst-served community, as the two policies serve different priorities. 

Moreover, the advantage of MAPPO over surplus sharing depends on the course of the event. It is largest when the disruption intensifies after onset, such as at delayed peak and progressive worsening. Learned coordination is therefore most valuable for prolonged or escalating events, whereas short, rapidly recovering events are handled well by simple rules.

Furthermore, independent learning is not a substitute for coordination. IPPO serves less of the network demand than every rule and has a lower worst-case service than MAPPO in every trajectory family. Centralized training needs network-wide records only during training, which can use historical or simulated data. At execution, each center still acts on its own information. 

Finally, redistribution comes at a sacrifice by the centers that give. In the Hurricane Ian case, the service of Lee County, represented by the surplus center of the network, falls from $0.918$ under local-only operation to $0.860$ under the sharing rules and to $0.766$ under MAPPO, which ships about $0.22$ of network demand compared with $0.13$ for uniform surplus sharing. Managers should set explicit limits on how far a donor's local service may fall, and should distinguish chronic shortages, which are better addressed through supply allocation and prepositioning before an event.

\section{Conclusion and Future Work}
\label{sec:conclusion}
In this study, we model decentralized redistribution among relief centers under partial environmental observation, with perishable inventory and delayed transportation, and solve the problem using recurrent MAPPO. We compare MAPPO with recurrent IPPO, local-only operation, and two decentralized surplus-sharing rules on held-out simulated disruptions and on a case study of Hurricane Ian.

With regard to RQ1, the recurrent MAPPO policy learns to balance local service, reserve, and redistribution without observing the underlying dynamics. It attains the highest mean worst-center service in four of the five hidden trajectory families and in the Hurricane Ian case, whose observed county-specific disruption trajectories were not used during training.

With regard to RQ2, coordination among centers improves service for the worst-served communities at the cost of overall network performance. In the Hurricane Ian case, MAPPO's worst-center service is significantly higher. The learned policy therefore trades total service for equity, and its advantage over simple rules is largest when the disruption persists or worsens.

With regard to RQ3, MAPPO significantly outperforms IPPO in network service and worst-center service. Because both learners use the same recurrent actor, this comparison identifies the benefit of centralized training rather than the incremental benefit of recurrence over a feedforward policy.

The study has limitations that suggest directions for future work. The case study maps outages and assistance registrations to demand and supply shocks by assumption. This illustrates policy behavior under a realistic disruption pattern rather than reproducing actual relief flows. The held-out tests draw new realizations of the five training families. Future work should also add transport and handling costs to the objective, evaluate the policies on recorded relief operations, and develop policies that transfer across network sizes.

\section*{Declaration of Generative AI}
During the preparation of this manuscript, the authors used Grammarly, ChatGPT, and Claude for language editing and grammatical review. The authors subsequently reviewed and edited the manuscript and take full responsibility for its content.

\appendix

\section{Synthetic Environment}
\label{app:generator}

Demand, available supply, and transportation conditions during an emergency are not known precisely in advance. Therefore, scenario-based formulations are commonly
used to represent uncertainty \citep{barbarosoǧlu2004two,chang2007scenario}. In this study, we construct a synthetic environment for our proposed learning model. We generate baseline demand and supply profiles that allow heterogeneity across relief centers, while stochastic shocks generate time-varying shortage, surplus, and balanced conditions that vary by the centers' roles.

\subsection{Hidden trajectories}
Each episode in the environment draws an onset day $\tau\in\{2,\dots,6\}$ and one hidden trajectory family $F$. The onset day $\tau$ is the first period in which the emergency event shock takes effect. We follow disaster studies to represent disruption using time-dependent functions, including simplified exponential and
trigonometric recovery paths \citep{bruneau2003framework,cimellaro2010framework}. Disaster studies also model uncertainty in the timing and dynamics of operational conditions
\citep{chang2007scenario,peng2014post}. We follow the relevant process and define five
trajectory families that represent early recovery, a delayed peak,
progressive worsening, persistent disruption, and aftershock. We set
$p=\frac{t-\tau}{T-\tau-1}$ and the recovery scale $\kappa=T$. The disaster intensity is $h_t=0$ before onset and $h_t=H_F(t;\tau)$ afterward.
The five trajectory families are defined by Eq.~\eqref{eq:families}.
\begin{subequations}
\label{eq:families}
\begin{align}
H_{\mathrm{ear}}
&=
\exp\left(
-\frac{t-\tau}{\omega_{\mathrm{ear}}\kappa}
\right),
\\[1pt]
H_{\mathrm{del}}
&=
\begin{cases}
p/\mu_{\mathrm{del}},
    & p\leq\mu_{\mathrm{del}},\\
\exp\left[-(p-\mu_{\mathrm{del}})/\omega_{\mathrm{del}}\right],
    & p>\mu_{\mathrm{del}},
\end{cases}
\\[1pt]
H_{\mathrm{wor}}
&=
b_{\mathrm{wor}}+
a_{\mathrm{wor}}p^{r_{\mathrm{wor}}},
\\[1pt]
H_{\mathrm{per}}
&=
b_{\mathrm{per}}+
a_{\mathrm{per}}\sin(2\pi p),
\\[1pt]
H_{\mathrm{aft}}
&=
\max\left\{
\exp\left(
-\frac{t-\tau}{\omega_{\mathrm{aft}}\kappa}
\right),
a_{\mathrm{aft}}
\exp\left[
-\frac{1}{2}
\left(
\frac{p-\mu_{\mathrm{aft}}}{\sigma_{\mathrm{aft}}}
\right)^2
\right]
\right\}.
\end{align}
\end{subequations}
All intensity values are clipped to $[0,1]$. The parameter values and the dynamics of the trajectory families used during training episodes are presented in
Table~\ref{tab:trajectory-parameters}. The trajectory parameters are selected to generate distinct temporal disruption patterns.

\begin{table}[H]
\centering
\caption{Hidden trajectory families and parameter values used in environment.}
\label{tab:trajectory-parameters}
\footnotesize
\setlength{\tabcolsep}{4pt}
\renewcommand{\arraystretch}{1.12}
\begin{tabularx}{\textwidth}{@{}p{3.0cm}p{5.2cm}X@{}}
\toprule
Family & Parameter values & Dynamics \\
\midrule
Early recovery
& $\omega_{\mathrm{ear}}=0.35$
& High initial disruption followed by gradual recovery. \\

Delayed peak
& $\mu_{\mathrm{del}}=0.55$, $\omega_{\mathrm{del}}=0.28$
& Disruption increases to a delayed peak and then declines. \\

Progressive worsening
& $b_{\mathrm{wor}}=0.15$, $a_{\mathrm{wor}}=0.85$,
  $r_{\mathrm{wor}}=1.4$
& Disruption becomes increasingly severe over time. \\

Persistent disruption
& $b_{\mathrm{per}}=0.72$, $a_{\mathrm{per}}=0.08$
& Disruption remains high with limited temporal variation. \\

Aftershock
& $\omega_{\mathrm{aft}}=0.25$, $a_{\mathrm{aft}}=0.90$,
  $\mu_{\mathrm{aft}}=0.72$, $\sigma_{\mathrm{aft}}=0.12$
& Initial disruption declines before a second disruption occurs. \\
\bottomrule
\end{tabularx}
\end{table}
These families generate distinct but unobserved temporal dynamics. Their differences motivate a recurrent policy that can infer the evolving state from its observation history, since the active family is not observed by the policy during training or evaluation.

\subsection{Demand and Supply Dynamics}
In addition to temporal uncertainty, the environment also introduces spatial
heterogeneity across relief centers, which is not provided explicitly to the actor or the critic.
At the beginning of each episode, we assign two centers randomly to a shortage role, two to a surplus role, and two to a balanced role. $r_i^{D}$ and $r_i^{S}$ determine the
baseline demand and supply conditions of center $i$, whereas $\beta_i$
and $\delta_i$ control the effects of disaster intensity on demand growth and
supply reduction, respectively. Additionally, $\eta_i$ controls the magnitude
of the delayed supply response, while $\chi_i$ determines transportation
vulnerability. 
\begin{table}[H]
\centering
\caption{Relief center-role parameter ranges.}
\label{tab:roles}
\footnotesize
\setlength{\tabcolsep}{3.2pt}
\renewcommand{\arraystretch}{1.12}
\begin{tabular}{@{}lcccccc@{}}
\toprule
Role & $r_i^{D}$ & $r_i^{S}$ & $\beta_i$ & $\delta_i$
     & $\eta_i$ & $\chi_i$ \\
\midrule
Shortage
& 1.05--1.20 & 0.80--0.95 & 0.40--0.90
& 0.35--0.70 & 0.10--0.40 & 0.30--0.75 \\
Surplus
& 0.85--0.95 & 1.10--1.30 & 0.00--0.15
& 0.00--0.15 & 0.30--0.70 & 0.05--0.25 \\
Balanced
& 0.95--1.05 & 0.95--1.05 & 0.10--0.35
& 0.10--0.30 & 0.10--0.35 & 0.10--0.40 \\
\bottomrule
\end{tabular}
\end{table}

The sampled roles with values in Table \ref{tab:roles} create a shortage center experiencing
higher demand, lower supply, and greater transport vulnerability, while surplus
centers experience the opposite, and balanced centers remain near baseline.

Given the sampled center roles and trajectory family, we next generate the
period-specific demand and supply at each center. Let
$h_t=H_F(t;\tau)$ denote the disaster intensity in period $t$, and let $z$
denote the episode-level severity multiplier. To capture a correlated network-wide
variation and local variation, $\epsilon_t^D$ and $\epsilon_{it}^D$ represent
common and center-specific demand disturbances, respectively. The resulting
demand at center $i$ is defined in Eq. \eqref{eq:demand}.
\begin{equation}
D_{it}
=
\bar D_i r_i^D
\left(1+z\beta_i h_t\right)
\exp\!\left(
\epsilon_t^D+\epsilon_{it}^D-\frac{1}{2}\sigma_D^2
\right),
\label{eq:demand}
\end{equation}
where $\bar D_i$ is baseline demand and $\sigma_D^2$ is the variance of the
center-specific demand disturbance $\epsilon_{it}^D$.

In the implementation, the episode severity is sampled as
$z\sim\mathrm{U}(0.80,1.20)$. The common and center-specific demand
disturbances are independent and sampled as
\[
\epsilon_t^D\sim\mathcal{N}(0,0.05^2),
\qquad
\epsilon_{it}^D\sim\mathcal{N}(0,0.08^2),
\]
with $\sigma_D=0.08$ in Eq.~\eqref{eq:demand}. The lognormal correction therefore removes the mean bias of the center-specific disturbance only, and the multiplicative demand noise has mean $\exp(0.05^2/2)\approx1.0013$ rather than exactly one. A correction for both disturbances would use the total variance $0.05^2+0.08^2=0.0089$; we report the model as implemented, since the resulting bias of about $0.1\%$ is negligible relative to the disruption effects.

Relief supplies do not increase immediately after the emergency event onset
\citep{AnayaArenas2014}. For each episode, a response delay
$d$ is sampled uniformly from $\{2,\dots,6\}$ days. After this delay, the supply
response initially increases and then gradually declines with Eq. \eqref{eq:supply-response}.
\begin{equation}
g_t=
\begin{cases}
0, & t<\tau+d,\\
\left(1-e^{-u/3}\right)e^{-u/(1.5\kappa)},
& t\geq\tau+d,
\end{cases}
\qquad
u=t-\tau-d.
\label{eq:supply-response}
\end{equation}

For the bootstrapped baseline supply $\widetilde S_{it}$, realized supply is
defined in Eq. \eqref{eq:supply}.
\begin{equation}
S_{it}
=
\widetilde S_{it}r_i^S
\max\left\{
0,\,
1-z\delta_i h_t+z\eta_i g_t
\right\}
\exp\!\left(
\epsilon_t^S+\epsilon_{it}^S-\frac{1}{2}\sigma_S^2
\right),
\label{eq:supply}
\end{equation}
where the common and center-specific supply disturbances are sampled as
\[
\epsilon_t^S\sim\mathcal{N}(0,0.04^2),
\qquad
\epsilon_{it}^S\sim\mathcal{N}(0,0.06^2),
\]
independently, and $\sigma_S=0.06$ is the standard deviation of the center-specific supply disturbance. As for demand, the correction applies to the center-specific term only, so the multiplicative supply noise has mean $\exp(0.04^2/2)\approx1.0008$; the total variance would be $0.04^2+0.06^2=0.0052$. The term $z\delta_i h_t$ represents the reduction in
supply caused by the emergency event, while $z\eta_i g_t$ represents the delayed
increase in incoming relief. The maximum operator ensures that the supply remains
nonnegative.

Table \ref{tab:simulator} lists the simulator settings that are not given in the equations above. 
\begin{table}[H]
\centering
\caption{Simulator settings (left) and center parameters (right). $n_i$ is the population weight of center $i$, $\bar D_i$ is its baseline daily demand (12\% assumed as population in need at 3.6 units per person), and $\bar S_i$ is the mean daily supply.}
\label{tab:simulator}
\footnotesize
\renewcommand{\arraystretch}{1.12}
\begin{minipage}[t]{0.56\textwidth}
\vspace{0pt}
\resizebox{\linewidth}{!}{%
\begin{tabular}{@{}ll@{}}
\toprule
Setting & Value\\
\midrule
Centers / horizon & $N=6$, $T=20$ days\\
Shelf life & $L=7$ days\\
Storage capacity & $K_i=5\,\bar D_i$\\
Outbound capacity & $\bar X_i=0.25\,\bar D_i$ per day\\
Initial inventory & $\min\{1.5\,\bar S_i,K_i\}$\\
Transport disruption & $\zeta^{\mathrm{trans}}_{it}=\min\{0.95,\,z\chi_i h_t\}$\\
Link disruption & $\bar\zeta^{\mathrm{trans}}_{ijt}=\max\{\zeta^{\mathrm{trans}}_{it},\zeta^{\mathrm{trans}}_{jt}\}$\\
Travel time & $\tau_{ijt}=1+\mathrm{Bin}\big(3,\min\{0.9,\bar\zeta^{\mathrm{trans}}_{ijt}\}\big)$\\
In-transit loss & $\rho_{ijt}\sim\mathrm{U}\big(0,0.2\,\bar\zeta^{\mathrm{trans}}_{ijt}\big)$\\
Reward & $w_1=2$, $w_2=3$, $w_3=4$, $\phi=0.60$\\
Scenarios & 160 / 20 / 20 (train / val. / test)\\
\bottomrule
\end{tabular}%
}
\end{minipage}\hfill
\begin{minipage}[t]{0.42\textwidth}
\vspace{0pt}
\resizebox{\linewidth}{!}{%
\begin{tabular}{@{}lrrrr@{}}
\toprule
Center & $n_i$ & $\bar D_i$ & $\bar S_i$ & $\bar S_i/\bar D_i$\\
\midrule
1 & 83{,}340 & 36{,}003 & 28{,}994 & 0.81\\
2 & 62{,}710 & 27{,}091 & 19{,}775 & 0.73\\
3 & 51{,}700 & 22{,}334 & 12{,}502 & 0.56\\
4 & 211{,}890 & 91{,}536 & 123{,}010 & 1.34\\
5 & 36{,}010 & 15{,}556 & 10{,}465 & 0.67\\
6 & 77{,}130 & 33{,}320 & 20{,}145 & 0.60\\
\midrule
Total & 522{,}780 & 225{,}841 & 214{,}890 & 0.95\\
\bottomrule
\end{tabular}%
}
\end{minipage}
\end{table}

\section{Hurricane Ian case-study construction}
\label{app:ian}
This appendix summarizes the data and the construction of the case-study environment in Section~\ref{sec:exp-ian}.

 Ian made landfall on Cayo Costa in Lee County on 28 September 2022 \citep{Bucci2023}. The case covers the 20 days from 26 September to 15 October 2022. We use (i) EAGLE-I county outage records at 15-minute intervals with the modeled number of electricity customers per county \citep{Brelsford2024}, (ii) OpenFEMA valid Individuals and Households Program registrations for DR-4673-FL by county \citep{OpenFEMA2026}. The six counties around the landfall designated for Individual Assistance each contributed one distribution point that operated in early October 2022 \citep{LeeCounty2022} shown in Table~\ref{tab:ian}. The daily outage share $o_{ct}$ for county $c$ is the mean number of customers without power over the 96 daily intervals, divided by the total number of customers. Center $i$ follows its own county path, $h_{it}=o_{c(i)t}/\max_{c,t}o_{ct}$, so that the most affected county (Charlotte, peak $0.844$) reaches $1$. The demand and supply effects are scaled by the impact factor $k_c=g_c/\max_{c'}g_{c'}$, where $g_c$ is the number of registrations per 1000 customers.

\begin{table}[H]
\centering
\caption{Hurricane Ian case study with counties, relief sites, and data. Customers are EAGLE-I modeled electricity customers; registrations are FEMA-valid Individuals and Households Program registrations; peak outage is the maximum daily share of customers without power between 26 September and 15 October 2022.}
\label{tab:ian}
\scriptsize
\setlength{\tabcolsep}{2.5pt}
\begin{tabularx}{\textwidth}{@{}l>{\raggedright\arraybackslash}Xrrrrr@{}}
\toprule
County & Point of distribution (address) & Customers & Registrations & per 1k & Peak outage & $k_c$\\
\midrule
Lee & Old Bonita Springs Library, 26876 Pine Ave., Bonita Springs & 628{,}635 & 193{,}944 & 308.5 & 0.648 & 0.725\\
Sarasota & Twin Lakes Park, 6700 Clark Rd., Sarasota & 310{,}646 & 57{,}363 & 184.7 & 0.657 & 0.434\\
Collier & Sugden Regional Park, 4284 Avalon Dr., Naples & 288{,}895 & 33{,}138 & 114.7 & 0.663 & 0.270\\
Manatee & Myakka City Community Center, 10060 Wauchula Rd., Myakka City & 266{,}408 & 28{,}477 & 106.9 & 0.485 & 0.251\\
Charlotte & Charlotte Sports Park, 2300 El Jobean Rd., Port Charlotte & 129{,}889 & 55{,}261 & 425.4 & 0.844 & 1.000\\
DeSoto & DeSoto County Emergency Management, 2200 NE Roan St., Arcadia & 21{,}129 & 8{,}122 & 384.4 & 0.736 & 0.904\\
\bottomrule
\end{tabularx}
\end{table}

\subsection{Hyperparameter Setting}
\begin{table}[H]
\centering
\caption{Learning configuration for the two policy-gradient methods (identical for both learners).}
\label{tab:hyper}
\footnotesize
\setlength{\tabcolsep}{3.5pt}
\renewcommand{\arraystretch}{1.12}
\begin{tabular}{@{}lcc@{}}
\toprule
Setting & Recurrent IPPO & Recurrent MAPPO\\
\midrule
Actor & GRU--128, Dirichlet & GRU--128, Dirichlet\\
Critic & local MLP--128--128 & CTDE MLP--128--128\\
Actor / critic learning rate & $3{\times}10^{-4}$ / $10^{-3}$ & $3{\times}10^{-4}$ / $10^{-3}$\\
Discount $\gamma$ / GAE $\lambda$ & 0.99 / 0.95 & 0.99 / 0.95\\
Clip $\epsilon$ / entropy coeff. $\eta$ & 0.20 / 0.01 & 0.20 / 0.01\\
Value-loss coeff. $c_V$ & 0.5 & 0.5\\
Epochs per rollout & 5 & 5\\
Episodes / transitions per update & 16 / 320 & 16 / 320\\
Recurrent minibatch & 4 episodes & 4 episodes\\
Episode length / training episodes & 20 / 30{,}000 & 20 / 30{,}000\\
\bottomrule
\end{tabular}
\end{table}
\clearpage

\bibliographystyle{elsarticle-harv}
\bibliography{cas-refs}
\end{document}